\documentclass[letterpaper, 10 pt, conference]{ieeeconf}  

\IEEEoverridecommandlockouts                              
\usepackage{graphicx}
\usepackage[]{amsmath}
\usepackage{amsfonts}
\usepackage{bm}
\usepackage{url} 

\usepackage{color}
\usepackage{soul}

\makeatletter
\let\MYcaption\@makecaption
\makeatother

\usepackage[font=scriptsize]{subcaption}
\makeatletter
\let\@makecaption\MYcaption
\makeatother

\title{\LARGE \bf
Motion Priority Optimization Framework toward Automated and Teleoperated Robot Cooperation in Industrial Recovery Scenarios
}

\author{Shunki Itadera and Yukiyasu Domae%
\thanks{The authors are with the Industrial Cyber-Physical Systems Research Center at National Institute of Advanced Industrial Science and Technology (AIST), Japan.
{\tt\small \{s.itadera, domae.yukiyasu\}@aist.go.jp}}%
}

\begin{document}
\maketitle
\thispagestyle{empty}
\pagestyle{empty}
\begin{abstract}
    
In this study, we introduce an optimization framework aimed at enhancing the efficiency of motion priority design in scenarios involving automated and teleoperated robots within an industrial recovery context.
The escalating utilization of industrial robots at manufacturing sites has been instrumental in mitigating human workload. 
Nevertheless, the challenge persists in achieving effective human-robot collaboration/cooperation (HRC) where human workers and robots share a workspace for collaborative tasks.
For instance, in the event of an industrial robot encountering a failure, such as dropping an assembling part, it necessitates the suspension of the corresponding factory cell for safe recovery. 
Given the limited capacity of pre-programmed robots to rectify such failures, human intervention becomes imperative, requiring entry into the robot workspace to address the dropped object while the robot system is halted. 
This non-continuous manufacturing process results in productivity loss.
Robotic teleoperation has emerged as a promising technology enabling human workers to undertake high-risk tasks remotely and safely. 
Our study advocates for the incorporation of robotic teleoperation in the recovery process during manufacturing failure scenarios, which is referred to as ``Cooperative Tele-Recovery''.
Our proposed approach involves the formulation of priority rules designed to facilitate collision avoidance between manufacturing and recovery robots. 
This, in turn, ensures a continuous manufacturing process with minimal production loss within a configurable risk limitation. 
We present a comprehensive motion priority optimization framework, encompassing an HRC simulator-based priority optimization and a cooperative multi-robot controller, to identify optimal parameters for the priority function.
The framework dynamically adjusts the allocation of motion priorities for manufacturing and recovery robots while adhering to predefined risk limitations. 
Through both quantitative and qualitative assessments, we validate the novelty of our concept and demonstrate its feasibility.
\end{abstract}

\section{Introduction}\label{sec:intro}
\begin{figure*}[!t]
    \centering
    \includegraphics[width=0.8\linewidth]{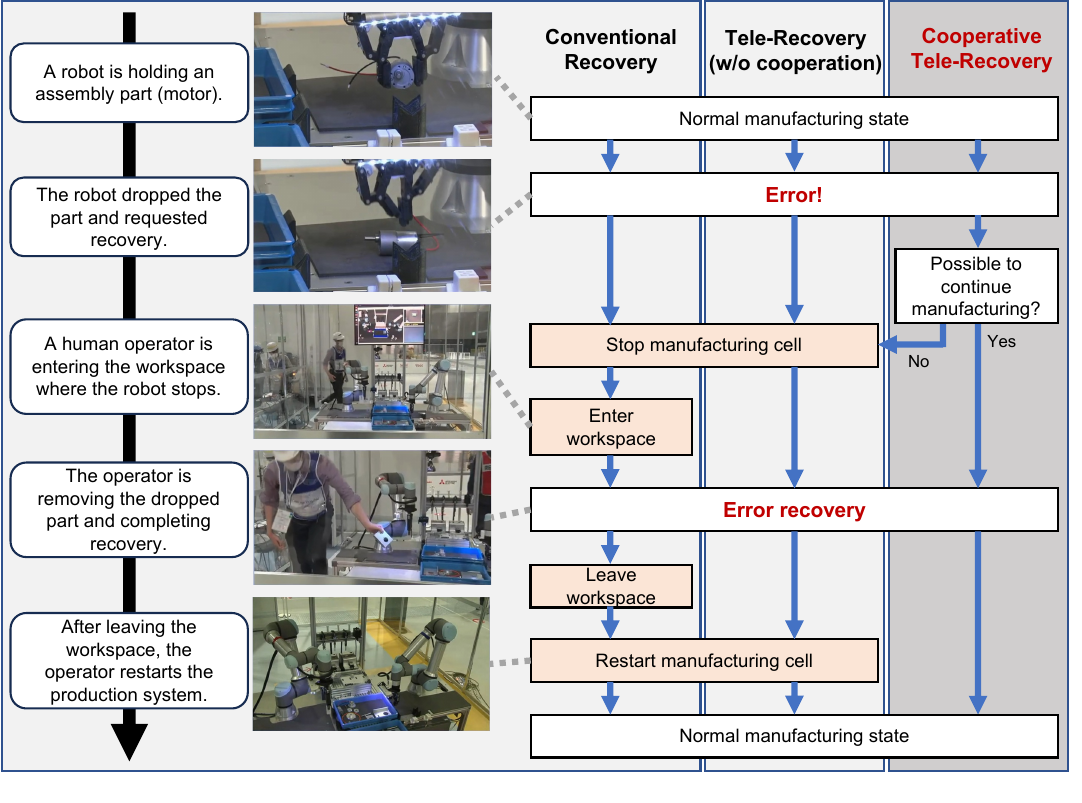}
    \caption{Our proposed concept of cooperative teleoperation recovery for industrial failure. The left snapshots are quoted from a video of World Robot Summit 2020, Assembly Challenge DAY2 (September 10, 2021)\protect\footnotemark[1] \cite{VonDrigalski2022}. The proposed approach ``Cooperative Tele-Recovery'' addresses eliminating the orange-colored processes, including stopping/restarting manufacturing robots and entering/leaving the shared workspace.}
    \label{fig:background}
\end{figure*}

The recent expansion of industrial robots has significantly contributed to the establishment of safe and efficient manufacturing processes, resulting in the development of autonomous factory systems \cite{IFR2022}. These robots have played a crucial role in supporting mass production within traditional industrial societies.
Recently, the potential of human-robot collaboration and cooperation (HRC) has been increasingly explored to enhance efficiency and productivity in shared workspaces with industrial robots \cite{Kopp2020}. HRC technology presents a promising solution aligning with the current trend of limited production of diversified products.
While mass production allows for dedicated system development, a versatile robotic system is required for variable-volume production systems.
Despite the introduction of numerous autonomous robots to industrial sites, their effective utilization in HRC poses persistent challenges. For instance, when a pre-programmed manufacturing robot encounters a failure, such as dropping a grasped object, it lacks the capability to autonomously address the failure. Consequently, human worker intervention becomes imperative for the recovery of the factory system.
However, a human operator cannot enter the workspace to address the failure until the production cell system comes to a halt for safety reasons, as illustrated in the left column "Conventional Recovery" of Fig.~\ref{fig:background}. In accordance with ISO\,10218-1 \cite{ISO10218-1}, a general recovery process typically involves the following steps:
\begin{enumerate}
    \item Stopping the whole or a part of the production cell.
    \item Entering the robot workspace.
    \item Recovering the production system, for example, removing the dropped part.
    \item Moving out of the workspace.
    \item Initializing and restarting the production system.
\end{enumerate}
This process is considered the most conservative approach in HRC to minimize collision risks between the human recovery worker and the autonomous manufacturing robot. ISO/TS\,15066 relaxes safety regulations in collaborative robot systems \cite{ISOTS15066}. However, the adoption of HRC work in industrial manufacturing sites remains limited due to the nascent state of HRC technology.
During the 2020 World Robot Summit, teams with robots unable to perform assembling tasks were prohibited from entering the workspace until their systems were deemed safe \cite{WRS2021,Yokokohji2022}. This practice leads to production suspension and potential deadlock situations, causing a reduction in the factory's productivity.
While advancements in robotic manufacturing technologies have been notable, with increasing success rates using sim-to-real technologies \cite{Narang2022,Tang2023}, practical implementation still requires improvement. Given the widespread use of manufacturing robots across numerous production cells in large factories, the possibility of failure recovery cannot be disregarded.
The prospect of simultaneous manufacturing and recovery processes becomes viable if innovative HRC technology sufficiently reduces collision avoidance risks. Enabling such concurrent operations contributes to minimizing productivity losses in industrial recovery scenarios.

This study explores the integration of robotic teleoperation technology into the human recovery process following manufacturing failures. Robotic teleoperation has garnered significant attention in industrial settings, primarily due to its potential to redefine factory job roles, allowing workers to operate remotely from any location worldwide. This paradigm shift holds promise for reducing the risk of work-related injuries and illnesses \cite{Zhang2023}.
Moreover, we anticipate that teleoperated robots can streamline the manufacturing recovery process. If teleoperated recovery robots are strategically positioned within each factory cell or can autonomously access the manufacturing workspace, a human operator can safely manage error recovery by directing the recovery robot. Consequently, teleoperation has the potential to eliminate the physical entry and exit processes associated with recovery, as depicted in the middle column ``Tele-Recovery (w/o cooperation)'' of Fig.\,\ref{fig:background}.
Nevertheless, this approach still necessitates the temporary suspension of the corresponding factory cell to prevent collisions between the manufacturing and recovery robots, resulting in reduced productivity. To address this, the reduction of productivity loss requires the omission of processes involving stopping and restarting the manufacturing system. In other words, the manufacturing process should persist even in the presence of minor abnormalities.
To achieve this, collision avoidance between manufacturing and teleoperation robots becomes instrumental, as illustrated in the right column ``Cooperative Tele-Recovery'' of Fig.\,\ref{fig:background}. In this Human-Robot Collaboration (HRC) approach, the production performance and associated risks, such as the number of assembly completions and the possibility of accidents due to dropped objects, are contingent on the collision avoidance policy. Adjusting this policy enables the factory system to manage performance and risk effectively, aiming to maximize productivity while adhering to acceptable risk limitations.
\footnotetext[1]{\url{https://www.youtube.com/watch?v=s9HmSVlMXao}}

\begin{figure}[!t]
    \centering
    \includegraphics[width=0.75\linewidth]{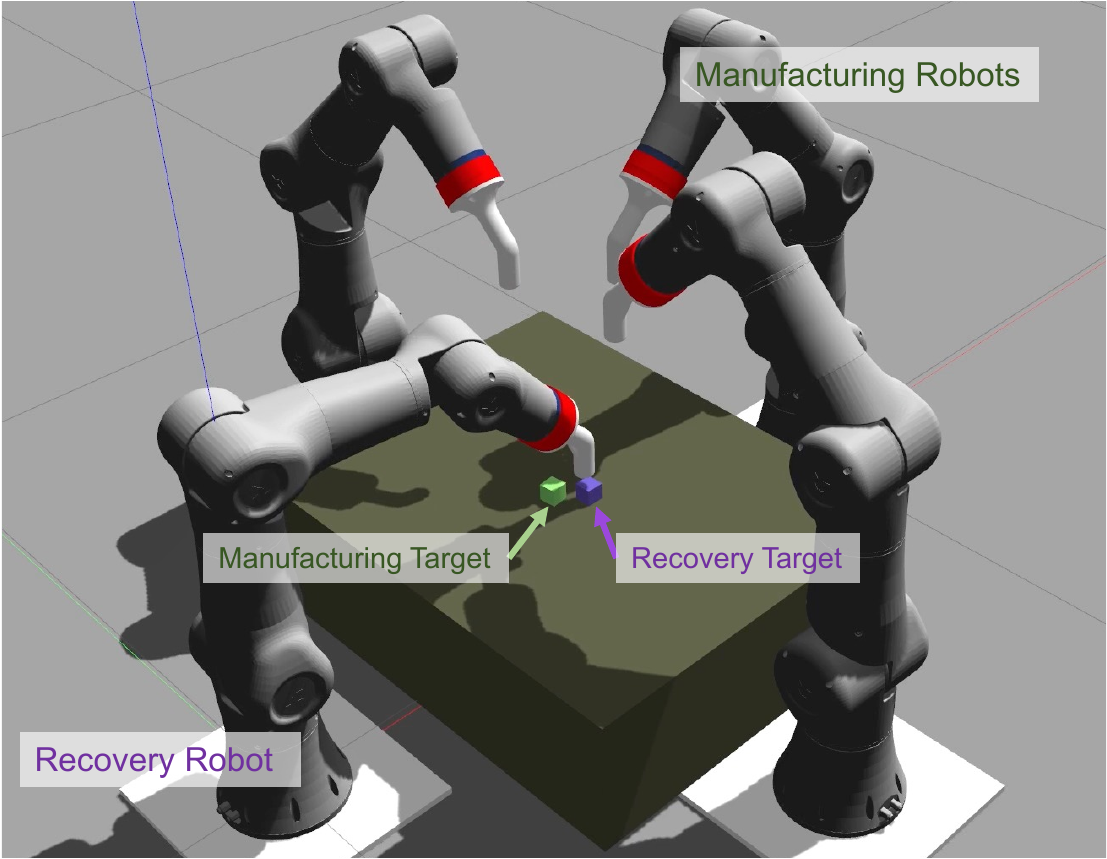}
    \caption{Proposed simulation environment designed for productivity-based motion priority optimization within a shared workspace. In the depicted scenario, the robot positioned in the foreground on the left is designated as a teleoperated recovery robot, while the others operate as autonomous manufacturing robots. To streamline the simulation, this study simplifies both recovery and manufacturing tasks, characterizing them as reaching tasks aimed at collecting datasets related to productivity and risk time.}
    \label{fig:recovery_situation}
\end{figure}

This study focuses on addressing HRC in an industrial setting with the aim of enabling a continuous manufacturing process and accomplishing recovery tasks with minimal production loss within an acceptable recovery time frame. Our approach involves the design of a motion priority function governing the interaction between manufacturing and recovery robots, incorporating a collision avoidance policy while balancing production and recovery requirements.
The proposed framework for the cooperative tele-recovery system comprises an HRC simulator for estimating productivity and recovery risk time, an optimization formulation to determine the optimal priority function using datasets from the simulator, and a cooperative multi-robot controller aligned with the priority function. Utilizing the HRC simulator enables the collection of datasets related to productivity and risk time in the recovery process, as depicted in Fig.\,\ref{fig:recovery_situation}.
Furthermore, we introduce an optimization method aimed at maximizing productivity, quantified as the number of tasks the manufacturing robot can complete while adhering to risk limitations. The risk limitation signifies the timeframe during which the recovery robot can address a dropped object. Relaxing this limitation constraint allows for the achievement of a recovery task with minimal production loss.
Through both quantitative and qualitative experiments, we substantiate the viability of our innovative concept of motion priority optimization in an industrial recovery scenario.

The rest of this paper is organized as follows:
Section \ref{sec:related_work} reviews the related studies.
Section \ref{sec:Problem_setting} presents our proposed framework.
Section \ref{sec:Implementation} describes the framework implemented in this study.
Section \ref{sec:hardware_exp} presents a preliminary experiment investigating the feasibility of
the presented optimization problem and the proposed collision avoidance algorithm with a robot hardware system.
Section \ref{sec:simulation_exp} presents an experiment where we conduct a qualitative assessment of the proposed framework.
Finally, Section \ref{sec:conclusion} concludes the study and discusses future work.

\section{Related Work}\label{sec:related_work}
\subsection{Autonomous Error Recovery}

While this study focuses on an HRC-based approach, extensive literature has also explored fully automated error-recovery processes. 
The error recovery process in automation can be divided into three steps as outlined in \cite{Loborg1994}: 1) detection, 2) diagnosis, and 3) recovery. 
The first two steps, which typically involve computer vision and do not require physical contact, have been widely researched. Recent applications of machine learning technologies have primarily focused on these steps, as reviewed in \cite{Kumar2023} and \cite{Park2020}. 
Notable examples include the use of the k-nearest neighbors algorithm and Bayes classifiers in semiconductor manufacturing processes \cite{Fan2020b} and sliding-window convolutional variational autoencoders for unsupervised anomaly detection in industrial robots \cite{Chen2020}. 
The use of explainable AI to understand and explain the causes of unexpected failures is discussed in \cite{Das2021}. 
Furthermore, several studies have designed comprehensive recovery plans incorporating all three steps, including a behavior-tree-based approach \cite{Wu2021}, a framework integrating computer vision with active vision and manipulation \cite{Kristiansen2020}, a conceptual graph method using case grammar and Bayesian networks \cite{Matsuoka2022}, and cost-oriented planning \cite{Nakamura2020}. 
However, while AI-based error detection and diagnosis are increasingly common in industrial settings, robot controllers for the recovery step, which involves interaction with the physical environment, are mostly confined to specific error situations. 
Therefore, human-worker cooperation remains crucial in the recovery process to handle diverse manufacturing failures.

\subsection{HRC in Industrial Scenarios}
The utilization of industrial robots, aimed at relieving workers from strenuous tasks and repetitive activities, has seen a notable increase.
However, human intervention remains essential due to the inherent limitations of robots, which may lack the capacity to think and adapt to unforeseen circumstances \cite{Vysocky2016}. 
Collaborative robots, also known as cooperative robots or cobots, are being actively developed not only to optimize overall system performance but also to mitigate the risk of occupational injuries and diseases.
The technical specification ISO/TS 15066 \cite{ISOTS15066} (Robots and robotic devices - Collaborative robots) serves as a comprehensive guide for collaborative robots, emphasizing safety in Human-Robot Collaboration (HRC). 
This critical aspect has spurred numerous studies dedicated to enhancing the security and reliability of robotic systems \cite{Baratta2022}. 
In this context, a range of sensors and algorithms has been introduced to ensure safety in industrial HRC scenarios \cite{Bonci2021}. 
Notable developments include the force/torque sensor-based safe admittance controller \cite{Mariotti2019} and a multi-camera-based safety barrier \cite{Ferraguti2020}.
Recently, there has been a surge in interest in cyber-physical systems (CPS) within industrial HRC. 
These systems play a crucial role in monitoring both the robot system and human workers, with the goal of achieving more efficient and secure production. 
This trend is evident in recent reviews \cite{Baratta2022,Pivoto2021}. 
Noteworthy studies have implemented CPS to assess real-time safety distances and provide closed-loop control for collision avoidance \cite{Nikolakis2019}. 
Additionally, CPS has been applied in task scheduling between human workers and autonomous robots to enhance manufacturing efficiency \cite{Maruyama2021}. 
Consequently, these CPS-based, or simulation-based, approaches hold considerable promise for estimating and enhancing the performance and safety of entire robotic systems in industrial settings.

\subsection{Robotic Teleoperation}
Robotic teleoperation, a technology enabling remote interaction with objects or environments, encompasses concepts such as telepresence \cite{Marvin1980} and telexistence \cite{tachi_book}. 
These concepts allow users to intuitively operate a robot and experience a sense of presence at a distant location. 
The applications of this technology span various fields, including industrial production \cite{Aschenbrenner2015}, space exploration \cite{Arzo2023}, construction \cite{Adami2021}, decommissioning \cite{Mizuno2023}, and medical care \cite{Mehrdad2021}. 
Different types of robots, such as industrial \cite{Gonzalez2021}, mobile \cite{Moniruzzaman2022}, and humanoid robots \cite{Darvish2023}, are utilized in these applications. 
Third-party technologies such as XR (AR, VR, and MR) \cite{Coronado2023} and 5G communication \cite{Moglia2022} have further augmented teleoperation capabilities.
A significant application of teleoperation lies in atypical work that is challenging to fully automate. 
Teleoperated robots leverage human knowledge and decision-making skills, making them more adaptable to dynamic environments than fully autonomous robots. 
This adaptability renders teleoperation particularly promising for industrial recovery scenarios, where operators can remotely manage recovery robots to address failures.
To the best of the authors' knowledge, no relevant studies on cooperative tele-recovery have yet been conducted.

One key challenge in teleoperation is semi-automated control, where both a human operator and an automated controller share authority over a robot or system. Shared control, a critical technology in semi-automation, combines remote operation with an intelligent assistant to enhance teleoperation performance and reduce operator burden \cite{Niemeyer2016}. Appropriate shifting of control authority is crucial for comfortable operation \cite{Abbink2012}. For example, prioritizing autonomous systems might limit intentional control by the operator, leading to frustration with unintended robot movements. Conversely, if the operator always has preferential control, the assistance system may not function effectively.
Recent studies have focused on developing algorithms for authority shifting and feedback interfaces, showing promising improvements in the performance and usability of teleoperated robots. Notable developments include a brain-machine interface for self-feeding in spinal cord injury patients \cite{Handelman2022}, driver behavior-based steering assistance \cite{Lu2023}, and intent prediction-based hydraulic manipulator teleoperation \cite{Luo2022}. In our target scenario, involving the collaboration of automated manufacturing robots and teleoperated recovery robots, an intelligent algorithm for collision avoidance is vital for safety. We posit that the concept of authority shifting could be key to achieving efficient cooperation between manufacturing and recovery robots. This paper focuses on the problem of robot priority design as a form of authority shifting in a Human-Robot Collaboration (HRC) recovery scenario.

\subsection{Collision Avoidance}
Collision avoidance in multi-robot systems has garnered significant attention in the realm of autonomous path planning, particularly in decentralized mobile robot control \cite{Raibail2022} and collaborative robotic manipulation \cite{Feng2020}. 
The primary objective is to plan collision-free paths to enhance the performance of autonomous robots. 
Algorithmic approaches include those based on localization uncertainty \cite{Hennes2012, Luo2020}, (deep) reinforcement learning \cite{Long2018,Fan2020a}, and model-based optimization \cite{Gafur2022}. 
These approaches often involve assigning distinct tasks to distributed, controllable robots, such as reaching specific target positions while avoiding collisions with other robots and static or dynamic obstacles.
Physical collision avoidance between autonomous robots and humans has also been explored in the context of safety in Human-Robot Collaboration (HRC) scenarios. 
Reviews on non-destructive disassembly \cite{Hjorth2022}, safety bounds \cite{Zacharaki2020}, and intuitive interfaces \cite{Villani2018} provide insights into this area. 
Notable examples include HRC architecture integrating trajectory adaptation and safety regulation \cite{Pupa2021}, an adaptive controller constraint based on the Explicit Reference Governor \cite{Merckaert2022}, and human-safety assessment-based trajectory scaling \cite{Lippi2021}. 
These studies typically treat humans as dynamic obstacles and plan the robot's path accordingly, considering estimations of human behavior. 
In this type of collision avoidance in HRC, robot trajectories are adaptively updated to ensure safety while accounting for robot performance.
However, traditional approaches often overlook optimizing the performance of the human worker. 
In our targeted HRC scenario, considering the performances of both automation and teleoperation robots (namely manufacturing and recovery robots, respectively) is crucial. 
This study addresses the challenge of designing motion priority for robots to maximize productivity by manufacturing robots while ensuring that the recovery duration by recovery robots does not exceed predefined risk limitations.

\section{Motion Priority Optimization Framework for Industrial Recovery Scenarios}\label{sec:Problem_setting}
\begin{figure}[t]
    \centering
    \includegraphics[width=0.9\linewidth]{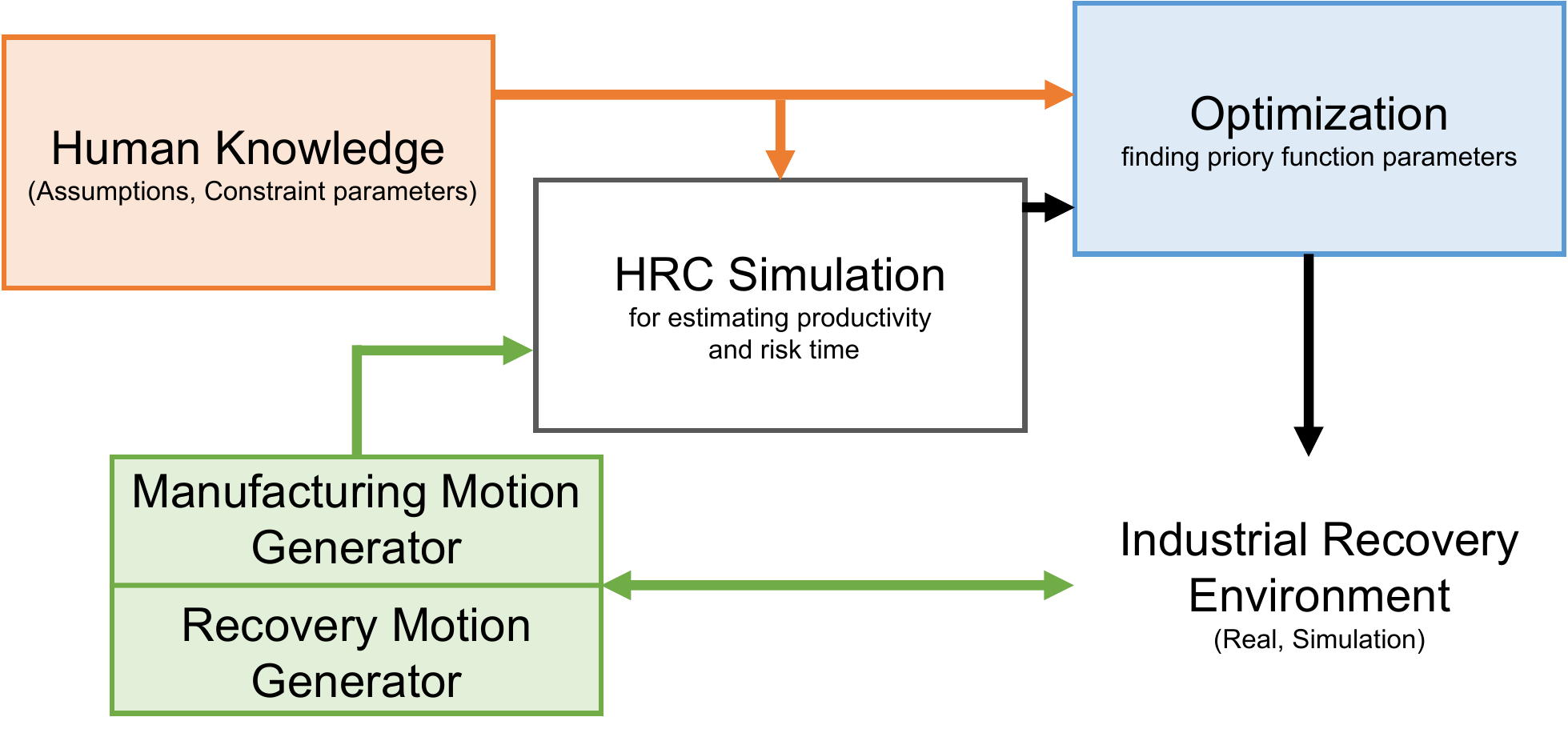}
    \caption{System structure of the proposed framework of productivity-based motion priority optimization, comprising 1) human prior knowledge (orange block), 2) manufacturing and recovery motion generators (green block), 3) HRC simulator (white block), 4) motion priority optimizer (blue block).}
    \label{fig:framework}
\end{figure}

This section introduces our proposed framework for cooperative tele-recovery in industrial scenarios, as shown in Fig.\,\ref{fig:framework}, which integrates components encompassing human knowledge, manufacturing and recovery motion generators, the HRC simulator, and an optimization formulation.

\subsection{Overview}\label{sec:overview}

This study aims to achieve HRC in an industrial recovery process with minimal productivity loss within acceptable risk limits. 
Our research concept aims to empower human operators to conduct recovery tasks without disrupting ongoing manufacturing processes. 
This functionality simplifies the recovery process and eliminates fatiguing tasks, such as physically entering the workspace. 
Additionally, compared with simply introducing a teleoperation robot for recovery, our framework enhances productivity by allowing the autonomous manufacturing process to continue. Our overarching research goal is to develop a framework that mitigates the negative impact of the recovery process on productivity.
The most conservative approach to motion priority design is to consistently prioritize the recovery robot over the manufacturing robot. 
This strategy is anticipated to minimize the risk time during which dropped objects exist in the workspace, ensuring the highest level of safety. 
However, this approach also results in a significant reduction in productivity. 
Alternatively, when dropped objects pose a low potential to impact the manufacturing process, minimizing risk time may not always be imperative.
In this study, we propose a method to relax the time constraint by introducing a recovery time limit, i.e., a risk time limit. 
This approach enables adaptive shifting of motion priority for productivity improvement. 
We develop a framework for deriving a priority function $p(*) \in [0,\ 1]$ that can continuously select between the manufacturing robot ($p=0$) and the recovery robot ($p=1$).

\subsection{Design Concepts and Structure}

\begin{figure}[!t]
    \centering
    \includegraphics[width=0.95\linewidth]{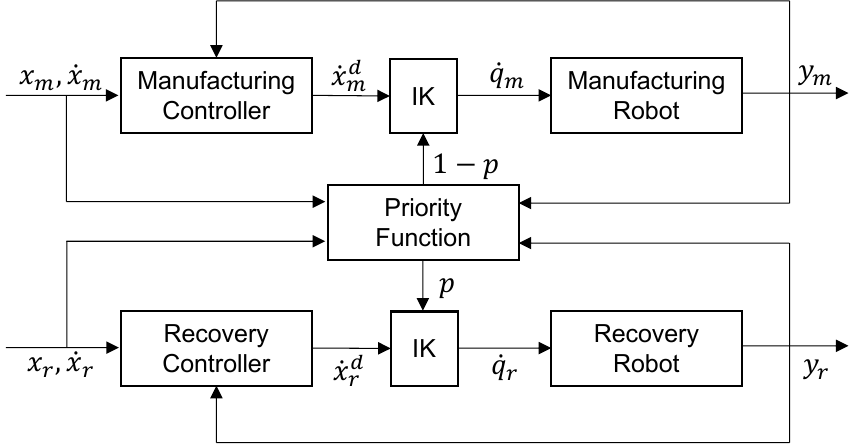}
    \caption{Block diagram of the proposed multi-robot control, composed of manufacturing and recovery robots, their controllers, IKs, and a priority function block.}
    \label{fig:block}
\end{figure}

This study develops the proposed framework to fulfill the following design objects:
\begin{itemize}
    \item Optimizing the motion priority function based on few-shot human demonstrations to reduce operator burden. 
    \item The error situation is considered in the form of a probabilistic distribution in the HRC simulator to fill in the gap between the simulator and the real environment.
    \item The system must be flexible for use in various robot controllers and priority functions, indicating a potential for future development (see Section \ref{sec:advantages}).
    \item A stable collision avoidance strategy is desired based on the optimized motion priority design. 
\end{itemize}
The framework structure depicted in Fig.~\ref{fig:framework} consists of four key components:
1) Human Prior Knowledge: This component (Component 1) provides task-specific target parameters and assumptions to other components based on the operator's experience and expertise. It includes information such as acceptable risk time and the prior distribution of dropped object positions.
2) Manufacturing and Recovery Motion Generators: Component 2 is responsible for producing the desired end-effector controllers for the manufacturing and recovery robots used in Component 3.
3) HRC Simulation: Component 3 is an HRC simulator used for collecting datasets related to the productivity and risk time of robots with randomly sampled priority functions.
4) Motion Priority Optimizer: Component 4, the optimizer, is designed to find the optimal priority function that maximizes productivity while adhering to the risk time limitation. This optimization is based on the datasets collected in Component 3.
It is noteworthy that these components are independent and can be implemented separately.
Collision avoidance, based on the optimal priority function, is implemented in a lower-level robot controller, such as inverse kinematics (IK). 
Figure \ref{fig:block} illustrates the block diagram representing our proposed framework, including manufacturing and recovery robots, their controllers, IKs, and a priority function block. 
The priority function block adjusts joint velocities in a colliding direction based on the target and current joint states. 
The suppression ratios, denoted as $p$ and $1-p$, are chosen from the range [0, 1], ensuring that they do not amplify robot motions. 
Therefore, if the robot controller is stable, the suppressed robot motion remains stable as well. 
Detailed formulations of the framework are presented in Section \ref{sec:Implementation}, along with an implementation example used for feasibility assessments in this study.

\subsection{Advantages of Our Framework}\label{sec:advantages}
Since the components of the framework can be implemented independently, the proposed framework offers the following advantages:

\subsubsection{Flexibility in the Motion Generators}

An inherent advantage of the proposed optimization framework lies in its flexibility within the robot controller. The robot motion generators in Component 2 are intentionally designed to be replaceable, operating independently from the collision-avoidance functionality embedded in lower-level controllers. This design choice ensures compatibility with a diverse array of controllers, spanning from model-based to learning-based controllers.
For instance, the framework seamlessly accommodates the utilization of an imitation learning algorithm to derive the recovery motion generator from human demonstrations. This incorporation allows the proposed simulation framework to discern an optimal priority that considers the distinctive features of human operations. This adaptability enhances the versatility of the optimization framework across various robotic control paradigms.

\subsubsection{Flexibility in the Priority Function Form}\label{sec:advantages_flex_priority}
Another notable advantage of the proposed framework is its flexibility in the form of the priority function. As the method presented is a data-based optimization approach, it accommodates various types of priority functions. This characteristic allows for the exploration and adaptation of different priority function forms based on specific requirements or characteristics of robot controllers. A future study is anticipated to delve into identifying appropriate priority function forms tailored to diverse robot controllers.
The independence of the framework components contributes significantly to its flexibility. If the components were tightly interdependent, the framework would be less adaptable and might require design as a singular comprehensive problem. While such an integrated system could potentially yield high-quality solutions, it might encounter challenges in accommodating more complex robot controllers and priority functions tailored for practical scenarios. Consequently, the proposed framework structure is viewed as a promising approach for practical applications, offering the adaptability necessary for addressing diverse and evolving robotic control challenges.

\subsection{Assumptions, Limitations, and Future Considerations}
\subsubsection{Controller Simplifications for HRC Simulator}
The proposed framework operates under the assumption that an HRC simulator can adequately capture the essential components of a realistic environment. In our current study, we need to simplify the robot motion controllers for both automated manufacturing and teleoperated recovery tasks in Component 2 to seamlessly integrate them into the HRC simulator. To extend the application of our proposed framework to an authentic factory setting, an exploration into more intricate motion controllers that better represent real-world scenarios may be necessary.
The emergence of photo-realistic simulators in recent developments \cite{Tang2023} presents a promising avenue for enhancing the accuracy of the simulation environment. This technological advancement could offer a valuable solution for refining the simulation realism within our framework. A future direction for our study involves an investigation into the validity of the motion controller simplifications made in our current system. This exploration aims to ensure that the implemented simplifications remain effective and applicable, contributing to the robustness and reliability of the proposed framework.

\subsubsection{Form of Priority Function}

A recognized limitation of the current framework is that, in practical applications, the form of the priority function $f(*)$ needs to be defined prior to data collection. The parameter sampling of this function is subsequently substituted for the priority function sampling in Component 3. Consequently, the performance of the optimization in Component 4 would be contingent upon the form of the predefined priority function.
A future study is envisioned to delve into identifying appropriate forms for the priority function tailored to various robot controllers. This investigation aims to enhance the adaptability and effectiveness of the proposed framework by exploring priority function structures that align optimally with different robotic control paradigms. By addressing this limitation, the framework can be further refined to cater to a broader spectrum of practical applications and scenarios.

\subsubsection{Human knowledge and Expertise}
Additionally, we acknowledge the potential challenge in estimating the acceptable recovery risk time and the prior distribution of dropped object positions in Component 1, as this may not always be straightforward for human operators. Given that our proposed framework represents a novel concept for the recovery process, further investigation is required to establish a reliable and effective approach for determining the risk time limitation.
Similarly, accurately estimating the prior distribution of dropped objects in real factory systems can be a complex task. With the ascent of digital transformation (DX) technologies, the collection of real-time failure data is becoming increasingly feasible, offering the potential to enhance the accuracy of our methodology. Therefore, we posit that our proposed framework holds significant potential for widespread application in future smart factories, particularly as advancements in technology facilitate improved data collection and decision-making processes.

\subsubsection{Advantage over Non-continuous Recovery Process}
The proposed continuous recovery process may not consistently outperform the conventional non-continuous approach. In situations where manufacturing and recovery tasks are relatively straightforward, and the rise in task execution time due to the continuous recovery process is not negligible, the non-continuous recovery process might prove to be more effective.
In practical applications, the proposed framework should include a mechanism to assess whether the continuous recovery process, based on the identified priority function, offers superior performance compared to the non-continuous counterpart. A detailed discussion on this aspect is presented in Appendix \ref{appendix:non-continuous}.

\section{Framework Implementations}\label{sec:Implementation}

\begin{figure*}[!t]
    \centering
    \includegraphics[width=0.8\linewidth]{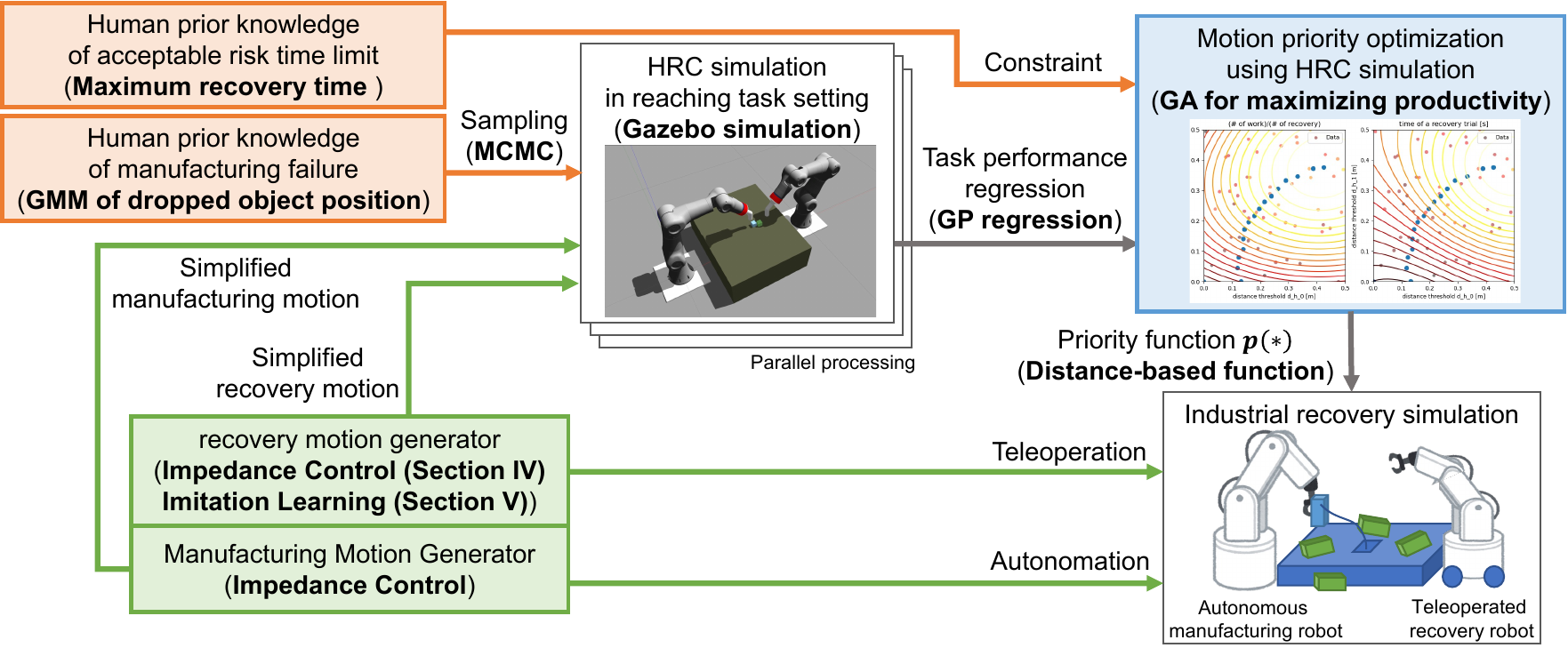}
    \caption{Implementation of the proposed framework. The color notations for components are consistent with those presented in Fig.~\ref{fig:framework}. To validate the viability of the proposed framework, we examine a simplified task employing versatile methodologies. The depictions in this figure presuppose a system consisting of one manufacturing robot and one recovery robot.}
    \label{fig:system}
\end{figure*}

This section presents the implementation of our proposed priority-optimization framework to underscore its feasibility. Figure \ref{fig:system} elucidates the implementation process used in this study. The framework and robot controllers are executed on a personal computer featuring an Intel Core i9-12900K CPU, 32 GB of RAM, and an Nvidia GeForce RTX 3070 GPU.

\subsection{Platform}
The platform employed in this study consists of either two (as detailed in Section \ref{sec:hardware_exp}) or four (as discussed in Section \ref{sec:simulation_exp}) robotic manipulators, specifically ToroboArm from Tokyo Robotics Inc., configured to face each other and share a common workspace. Among these robots, one is designated for teleoperated recovery, while the remaining units are utilized for autonomous manufacturing. Each robot possesses seven degrees of freedom (DoF) and is equipped with a velocity controller for each joint. At the end effectors, six-axis force/torque sensors are installed to capture contact forces. The controllers are seamlessly integrated into the Robot Operating System (ROS) Noetic on Ubuntu 20.04. The system operates at a frequency of 500 Hz.
It should be noted that the compatibility of the proposed framework is not limited to these platform configurations.

\subsection{Low-level Robot Controllers}

This section introduces a low-level controller that incorporates IK with collision avoidance for both manufacturing and recovery robots. The outlined low-level control framework has been implemented within our developing open-source library, \textit{OpenHRC}\footnote[2]{\url{https://github.com/itadera/OpenHRC}}.
For future scalability, we employ $N_m$ and $N_r$ to signify the number of autonomous manufacturing and teleoperated recovery robots, respectively. The aggregate number of robots is denoted as $N = N_m + N_r$. Here, $n_i$ represents the DoF, and $\dot{\bm{q}_i}$ denotes the joint velocity vector of the $i$-th robot, where $i = 1, 2, \cdots, N$ serves as the index for all robots. Additionally, we utilize $i_m=1, 2, \cdots, N_m$ and $i_r=1, 2, \cdots, N_r$ as indices for the manufacturing and recovery robots, respectively.

An IK with collision avoidance can be formulated as the following optimization problem in the framework of quadratic programming (QP):
\begin{align}
    \begin{array}{rl}
        \mathop{\rm arg~min}\limits_{\dot{\bm{q}}} & \displaystyle\sum_i^N w_i \Arrowvert \hat{\bm{J}}_i(\bm{q}) \cdot \dot{\bm{q}} - \dot{\bm{x}}_i \Arrowvert^2 +\varepsilon \Arrowvert \dot{\bm{q}} \Arrowvert^2 \\ [+10pt]
        \mbox{s.t. }                               & \dot{\bm{q}}_{min} \le \dot{\bm{q}} \le \dot{\bm{q}}_{max}                                                                                                     \\
                                                   & b < \bm{A} \cdot \dot{\bm{q}}
    \end{array}
    \label{eq:ik}
\end{align}
where $\bm{q} = [\bm{q}_1^T\ \bm{q}_2^T\ \cdots\ \bm{q}_N^T]^T \in \mathbb{R}^{\sum_N n_i}$ and $\dot{\bm{q}} = [\dot{\bm{q}}_1^T\ \dot{\bm{q}}_2^T\ \cdots\ \dot{\bm{q}}_N^T]^T \in \mathbb{R}^{\sum_N n_i}$ are concatenated vectors of all robot joint angles $\bm{q}_* \in \mathbb{R}^{n_*}$ and velocities $\dot{\bm{q}}_* \in \mathbb{R}^{n_*}$.
$\hat{\bm{J}}_i(\bm{q}) = [\bm{0} | \bm{J}_i(\bm{q}_i) | \bm{0}] \in \mathbb{R}^{6 \times \sum_N n_i}$ is an augmented sparse matrix of the same size as the horizontally concatenated matrix of all robot Jacobians, but $i$-th robot Jacobi matrix $\bm{J}_i(\bm{q}_i)$ is substituted for the corresponding part of the augmented matrix.
$\dot{\bm{x}}_i = [\dot{\bm{p}}^T_i,\ \bm{\omega}^T_i]^T \in  \mathbb{R}^6$ is the desired cartesian velocity vector of the $i$-th robot end effector, where $\dot{\bm{p}}_i$ and $\bm{\omega}_i$ are the translational and angular velocities, respectively.

The objective function in eq.~\eqref{eq:ik} consists of the following two terms:
The first term aims to minimize the distance between the robot end-effector velocity and the target velocity. The weight $w_i$ corresponds to the priority level of the robot in collision avoidance, as further elucidated below.
The second term serves to stabilize the robot joint velocity, particularly in the singularity posture, acting as damping with weight $\varepsilon$.

\begin{figure}
    \centering
    \includegraphics[width=0.95\linewidth]{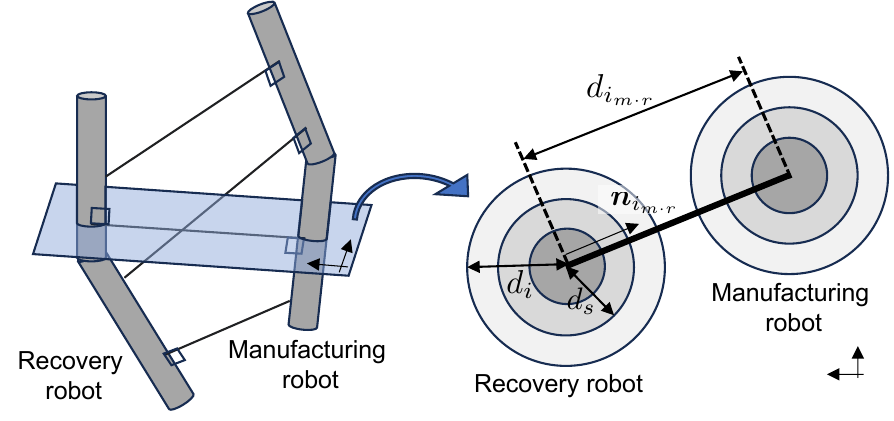}
    \caption{Illustration of the collision avoidance constraints in eq.\,(\ref{eq:collision_avoidance}). This figure depicts a scenario where two two-link robots are in collision, and four combinations of the robots' links are considered. In our practical implementation, constraints are formulated for all possible combinations of the robots' links. Consequently, at maximum, ${}_{\Sigma (n_i+1)} \mathrm{C}_2$ constraints are added to the QP-based IK optimization problem.}
    \label{fig:collision_avoidance_ik}
\end{figure}

\begin{figure}[t]
    \begin{center}
        \begin{subfigure}[t]{0.32\linewidth}
            \centering
            \includegraphics[clip,width=0.99\linewidth]{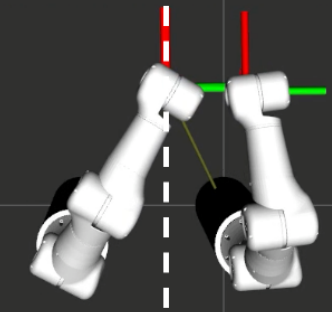}
            \caption{Prioritizing left robot.}
        \end{subfigure}
        \begin{subfigure}[t]{0.32\linewidth}
            \centering
            \includegraphics[clip,width=0.99\linewidth]{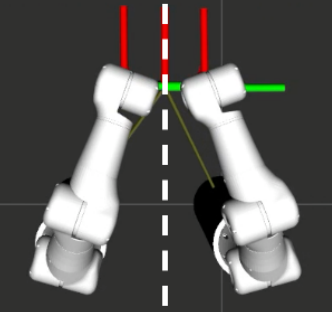}
            \caption{Prioritizing equally.}
        \end{subfigure}
        \begin{subfigure}[t]{0.32\linewidth}
            \centering
            \includegraphics[clip,width=0.99\linewidth]{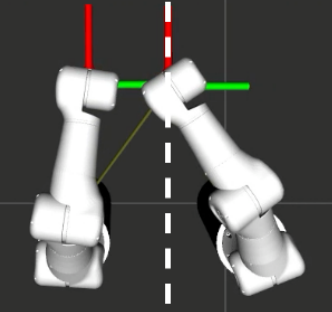}
            \caption{Prioritizing right robot.}
        \end{subfigure}
        \caption{Examples of motion prioritization by significantly increasing a weight $w_i$ in eq.~(\ref{eq:ik}) corresponding to (a) the left robot and (c) the right robot. The white broken lines indicate the target position of the robots. (b) Illustration of the case of setting the same weights, where neither robot can reach the target position.}
        \label{fig:prioritize_test}
    \end{center}
\end{figure}

This optimization problem incorporates two types of constraints. The first constraint ensures that the joint positions and velocities remain within predefined limits, where $\dot{\bm{q}}_{min}$ and $\dot{\bm{q}}_{max}$ represent the minimum and maximum joint velocity, respectively. We introduce the second constraint for collision avoidance, drawing on pertinent studies that address Jacobian-based collision avoidance between a high-DoF robot body and convex-shaped objects \cite{Faverjon2006, Kanehiro2009}. To enforce collision avoidance between the $i_m$-th and $i_r$-th robots, we formulate a constraint equation in terms of the relative distances.
Specifically, we represent the constraint using the following equations:
\begin{align}
    \begin{aligned}
        \bm{A} & = \bm{n}_{i_{m\cdot r}}^T\cdot (\hat{\bm{J}}_{i_m} - \hat{\bm{J}}_{i_r}), \\
        b      & = -\xi \frac{d_{i_{m\cdot r}}-d_s}{d_i - d_s},       
    \end{aligned}
    \label{eq:collision_avoidance}
\end{align}
where $d_{i_{m\cdot r}} = \Arrowvert\bm{p}_m - \bm{p}_r \Arrowvert^2$ is the relative distance between end effectors.
$\bm{n}_{i_{m\cdot r}} = (\bm{p}_m - \bm{p}_r)/ d_{i_{m\cdot r}}$ denotes a normal unit vector from $\bm{p}_r$ to $\bm{p}_m$.
$\xi$ is a gain for convergence speed adjustment.
$d_i$ and $d_s$ are denoted as influenced distance and security distance, respectively, as shown in Fig.\,\ref{fig:collision_avoidance_ik}.
In our implementation, we assume that the shapes of the robot links are cylindrical. When the relative distance between the end effectors is less than the influenced distance, i.e., $d_{i_{m\cdot r}} < d_i$, the constraint suppresses the robot velocity in the approaching direction with respect to the relative distance $d_{i_{m\cdot r}}$. Upon reaching the security distance, i.e., $d_{i_{a-t}} = d_s$, the relative velocity in the approaching direction is regulated to zero. Consequently, the constraint effectively hinders the robots from moving closer and mitigates collision risks.
The collision avoidance approach constrains the relative distances between the robots. Therefore, the weight $w_*$ in the objective function in eq.~\eqref{eq:ik} determines the control priority of the robots. Figure \ref{fig:prioritize_test} illustrates three weight conditions for two colliding robots. When the weight is significantly larger than the other weights, the motion of the corresponding robot takes precedence. The design method for determining the weight $w_i$ is expounded in Section \ref{sec:priotiry_function}.
The QP form is widely employed in numerical optimization due to its efficient solving methods. In our current implementation, we utilize the QP solver \textit{OSQP} as proposed in \cite{Stellato2020}.

\subsection{Priority Function}\label{sec:priotiry_function}

Our framework offers the capability to prioritize one robot over another by significantly increasing the weight $w$ in eq.~\eqref{eq:ik} associated with the specific robot, thereby facilitating the switching of authority between the robots. To mitigate the risk of a robot getting stuck due to conflicts in priority and the system reaching a deadlock, we employ a simple function for the priority function. As outlined in Section \ref{sec:overview}, $p=0$ and $p=1$ correspond to prioritizing manufacturing and recovery robots, respectively.
Various function arguments and forms can serve as candidates for the priority function. To streamline the problem within a heuristic approach, we utilize relative distances $l_{i_m}$ between the end effector of the ${i_m}$-th autonomous manufacturing robot and the current target object as function arguments. The priority function is represented as follows:
\begin{equation}
    p(l_{i_m} | \bar{l}_{i_m}) = \begin{cases}
        0 \ \ \mathrm{if}\ l_{i_m} < \bar{l}_{i_m}, \\
        1   \ \ \mathrm{otherwise},                   \\
    \end{cases}
    \label{eq:priotiry}
\end{equation}
where $\bar{l}_{i_m}$ is a threshold determining when to switch priority.
When the relative distance $l_{i_m}$ is smaller than threshold $\bar{l}_{i_m}$, the proposed framework prioritizes the autonomous manufacturing robot.
Otherwise, the teleoperation recovery robot is prioritized over the autonomous robots.
We can transform the design problem of priority function $p(*)$ into an optimization problem of the thresholds $\bar{\bm{l}} = [\bar{l}_{1} \cdots \bar{l}_{N_m}]^T $.

The priority function in eq.~(\ref{eq:priotiry}) does not incorporate any terms related to the teleoperated recovery robot. Consequently, priority switching remains unaffected by the human operator's behavior, ensuring the safety of the proposed framework in the event of risky operation input from the operator. In the subsequent sections, we operate under the assumption that the operator's input does not involve any unexpected operations.
Utilizing the output of the priority function, we adjust the weights in eq.~(\ref{eq:ik}) corresponding to the robot priority in the recovery process as follows.
\begin{align}
    \begin{aligned}
        w_{i_r} & =  w_o \cdot \gamma^{p},   \\
        w_{i_m} & =  w_o \cdot \gamma^{1-p},
    \end{aligned}
\end{align}
where $w_o$ is a nominal weight and $\gamma$ is a coefficient for significantly increasing the weights.

\subsection{High-level Robot Controllers}\label{sec:high-level-controller}

In this section, we introduce simplified high-level controllers for autonomous and teleoperated robots, designed to generate desired end-effector velocities for simulating HRC manufacturing and recovery tasks. The control framework is embedded within our developing library, \textit{OpenHRC}. The high-level controller employed for both manufacturing and recovery movements is rooted in an impedance control framework.
Furthermore, the recovery motion controller, as demonstrated in a qualitative experiment in Section \ref{sec:simulation_exp}, incorporates probabilistic movement primitives. These primitives are utilized to provide a human-like target trajectory, serving as the desired positions and velocities within the impedance controller.

\begin{figure}
    \centering
    \includegraphics[width=0.7\linewidth]{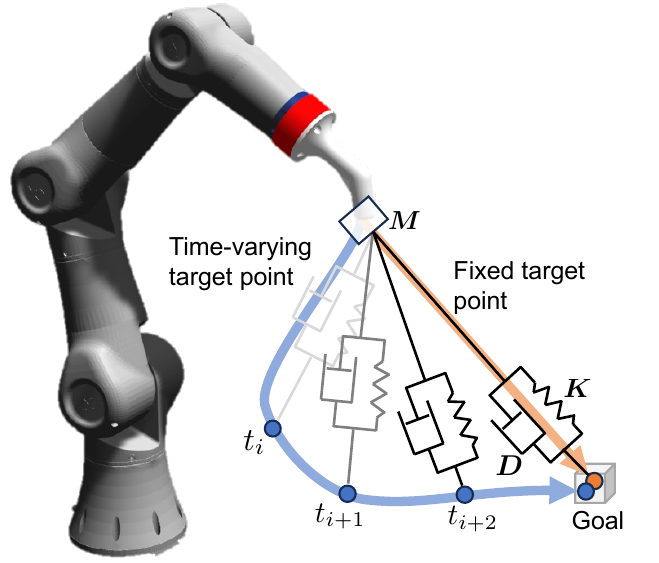}
    \caption{Impedance controller whose target ($\bm{p}$ and $\dot{\bm{p}}$) are fixed to the goal and following a time-varying trajectory, illustrated as orange and blue lines, respectively.}
    \label{fig:high-level-controllers}
\end{figure}

\subsubsection{Impedance Controller}
Impedance control is a widely adopted approach for implementing compliant robot control. This method is characterized by its passivity and robustness against disturbances, ensuring convergence to the target.
The control formulation for a robot is
\begin{align}
    \begin{aligned}
        \bm{M} \ddot{\bm{p}} + \bm{D} (\dot{\bm{p}} - \dot{\bm{p}}^d) + \bm{K}(\bm{p} -\bm{p}^d) & = \bm{f}, \\
    \end{aligned}
    \label{eq:ic}
\end{align}
where $\bm{p}^d$ and $\dot{\bm{p}}^d$ are the desired position and velocity vectors of a robot end effector.
$\bm{M} \in \mathbb{R}^{3\times 3}$, $\bm{D} \in \mathbb{R}^{3\times 3}$ and $\bm{K} \in \mathbb{R}^{3\times 3}$ are diagonal matrices comprising the virtual horizontal mass, damper, and spring coefficients in three dimensions, respectively.
$ \bm{f}\in \mathbb{R}^3$ is an external force vector applied to the robot end effector.
By applying Euler Integration to eq.~(\ref{eq:ic}), we obtain the cartesian velocity command $\dot{\bm{p}}$ of the \textit{i-}th robot end-effector.

The impedance controller facilitates the continuous approach of the robot end-effector to the target pose while remaining robust to disturbances, including collision avoidance. To stabilize the motions of the robot end-effectors, we design the damping coefficients to achieve critical damping motion, characterized by $d = 2 \sqrt{ mk}$, where $m$, $d$, and $k$ represent the diagonal elements of $\bm{M}$, $\bm{D}$, and $\bm{K}$, respectively.
Adjusting the virtual coefficients allows for the modification of movement speed and handling time, enabling the simulation of various manufacturing and recovery tasks, such as picking and assembling. Throughout this paper, with the exception of a qualitative experiment in Section \ref{sec:simulation_exp}, we employ fixed points as targets. In this context, $\bm{p}^d$ remains a constant vector, and $\dot{\bm{p}}^d$ is a zero vector, as depicted by the orange line in Fig.~\ref{fig:high-level-controllers}.

\subsubsection{Probabilistic Movement Primitives}\label{sec:promp}
To incorporate human behavior into the Human-Robot Collaboration (HRC) simulator, we utilize imitation learning, allowing the generation of a human-like trajectory based on human demonstrations. In the qualitative experiment outlined in Section \ref{sec:simulation_exp}, we leverage Probabilistic Movement Primitives (ProMPs) to render $\bm{p}^d$ and $\dot{\bm{p}}^d$ as time-dependent functions, thereby replicating human teleoperation within the developed simulation.
ProMPs represent a popular imitation learning method capable of modeling time-varying trajectory variance, thereby capturing multiple human demonstrations \cite{Paraschos2013}. It has been widely employed for reproducing human dynamic movements \cite{Gomez-Gonzalez2020}. This approach serves as a practical example of integrating human recovery operations into the proposed HRC simulator.
Briefly, the generated target position and velocity at each time step are represented as follows:
\begin{align}
    \begin{aligned}
        \bm{y}^d = \begin{bmatrix}
            \bm{p}^d  \\
            \dot{\bm{p}}^d 
            \end{bmatrix} = \bm{\Phi} \bm{w} + \bm{\epsilon},\\
            p(\bm{y}^d|\bm{w}) =  \mathcal{N}(\bm{y}^d | \bm{\Phi} \bm{w}, \bm{\Sigma}),
    \end{aligned}
    \label{eq:promp}
\end{align}
where $\bm{w}$ is a learnable weight vector, $\bm{\Phi}$ is a basis function matrix, and $\bm{\epsilon} \sim \mathcal{N}(\bm{0}, \bm{\Sigma})$ is a zero mean i.i.d Gaussian noise with uncertainty $\bm{\Sigma}$.
In this study, we utilize ProMPs implementation \textit{Movement Primitives}\footnote[3]{\url{https://github.com/dfki-ric/movement_primitives}} to generate the target position and velocity for the recovery robot.
By employing a time-varying target point for both the generated trajectory's position, denoted as $\bm{p}$, and velocity, represented by $\dot{\bm{p}}$ in eq.\,(\ref{eq:ic}) at each time step, the controller demonstrates the ability to produce human-like end-effector movements with a notable resilience against disturbances. This is depicted by the blue line in Fig.~\ref{fig:high-level-controllers}.
It should be noted that in this study, the target trajectory is not regenerated during a trial to ensure the smoothness of the recovery movement, even if the other robots encumber the recovery robot.
Therefore, the recovery motion generator handles the end-effector to follow the initially generated trajectory, which may be an unnatural human motion in some cases.
Future work will consider the restoration movement of the recovery robot after collision avoidance.

\begin{figure}[!t]
    \centering
    \begin{minipage}[b]{0.55\hsize}
        \centering
        \includegraphics[clip,width=0.99\linewidth]{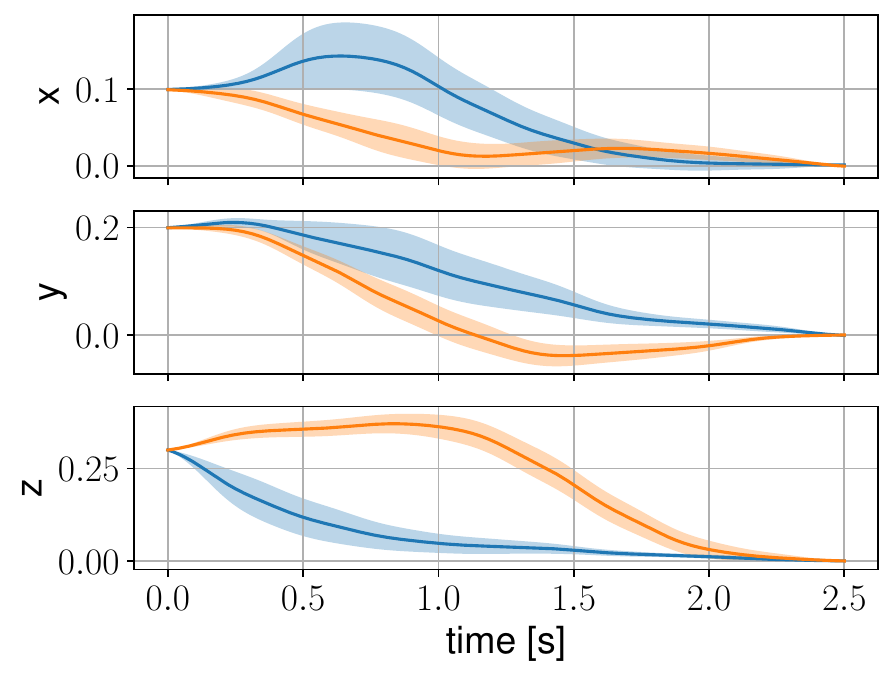}
    \end{minipage}
    \begin{minipage}[b]{0.42\hsize}
        \centering
        \includegraphics[clip,width=0.99\linewidth]{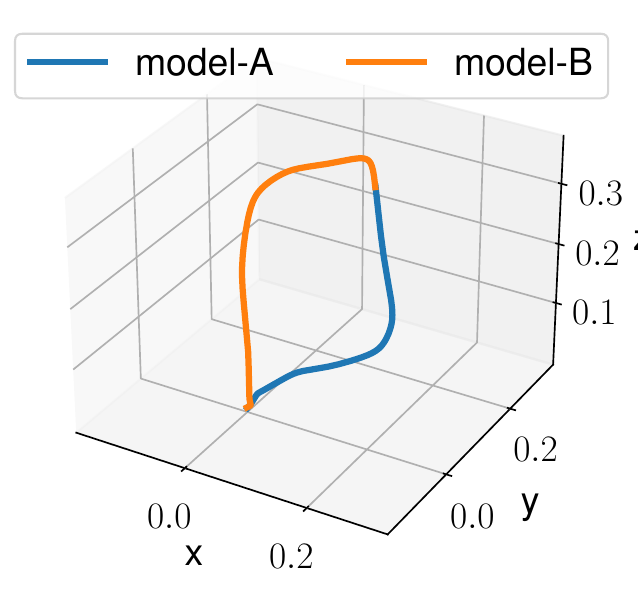}
    \end{minipage}
    \caption{Examples of target trajectories generated by the trained ProMPs models (model-\{A, B\}). The start and goal positions are $[0.1,0.2,0.3]$ and the origin, respectively. The operation duration is 2.5\,s. The training dataset for each model is described in Appendix \ref{appendix:data-collection}.}
    \label{fig:ProMPs}
\end{figure}

This study utilizes two ProMPs models, each trained with distinct trajectory types, serving to encapsulate a spectrum of human behaviors. The specifics regarding the training datasets are elucidated in Appendix \ref{appendix:data-collection}.
After training the ProMPs models with the respectively generated datasets, the controllers exhibit the capacity to generate diverse trajectories even when commencing from identical start point and concluding at the same goal positions. Figure \ref{fig:ProMPs} visually represents trajectories generated by the trained ProMPs models (model-\{A, B\}) with the start position $\bm{p}_0 = [0.1,0.2,0.3]$ and the goal position $\bm{p}_g = \bm{0}$.
These trained models adeptly capture the distinctive features inherent in each recovery motion dataset.

\subsection{Data Sampling}
The proposed framework encompasses two data sampling processes designed to emulate a human knowledge-based HRC and determine the optimal thresholds denoted as $\bar{\bm{l}}$.

\subsubsection{Prior Sampling for HRC simulation}
\begin{figure}[!t]
    \centering
    \includegraphics[width=0.95\linewidth]{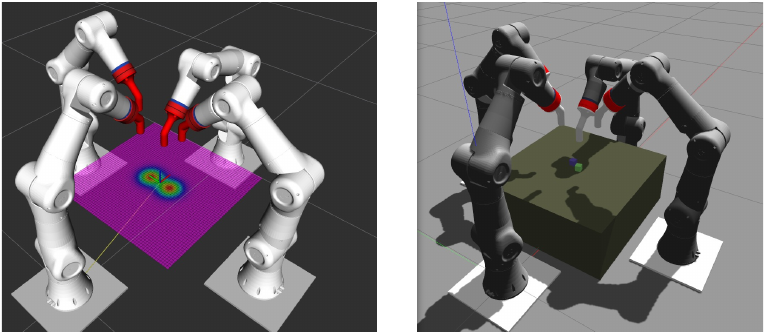}
    \caption{Probabilistic distribution of the recovery objects. 
In this study, we posit that objects manifest on both the left and right sides of the fixed manufacturing object, depicted as a coordinate axis, as perceived from the viewpoint of the recovery robot. This distribution adheres to a predefined mixture Gaussian distribution. }
    \label{fig:prior_distribution}
\end{figure}

In the context of simulating a simplified HRC, error data is generated in the manufacturing process, specifically pertaining to the positions of dropped objects. However, acquiring raw error data in real manufacturing settings poses challenges. To address this, our study makes the assumption that the probabilistic distribution of object dropping is more prevalent around the left and right sides of a fixed manufacturing point, as perceived from the recovery robot's perspective.
This prior distribution is modeled as a Gaussian mixture model, as depicted in Fig.~\ref{fig:prior_distribution}. To generate data samples from this prior distribution, we employ the Markov Chain Monte Carlo method, chosen for its flexibility in handling a broad spectrum of prior distributions. The sampling process is implemented using the open-source library \textit{PyMC} \cite{Salvatier2016}.
Upon reaching the target position, the recovery robot simulator relocates the target recovery object based on the sampled data from the prior distribution. This process is iteratively carried out to simulate diverse scenarios.

\subsubsection{Simulation-based HRC Sampling}
To generate simulation data for HRC, a Gazebo-based simulator has been developed, facilitating concurrent manufacturing and recovery processes. The controllers and data sampling methods described earlier are employed within this simulator. In Fig.~\ref{fig:recovery_situation}, the recovery robot is directed to reach a purple target box, while the manufacturing robots are oriented to achieve a green target box through rotation.
The placement of recovery target boxes within the Gazebo simulator is executed using the \textit{pcg\_gazebo} Python library\footnote[4]{\url{https://boschresearch.github.io/pcg_gazebo/}}. By utilizing this simulator and randomly selecting thresholds $\bar{\bm{l}} \sim \mathcal{U}(0, l_{max})$ in the priority function (eq.~(\ref{eq:priotiry})), data is gathered on the estimated productivity and recovery risk. Specifically, this encompasses the number of manufacturing tasks achievable during one recovery task, denoted as $X_{product}$, and the duration required to complete the recovery process, denoted as $X_{risk}$.
The parameter distribution $\mathcal{U}(0, l_{max})$ signifies a uniform distribution ranging from 0 to $l_{max}$, with this study adopting $l_{max} = 0.5$. The HRC sampling outcomes are contingent on the chosen thresholds $\bar{\bm{l}}$ due to the interactive nature of the robots.
In instances where deadlock situations arise, impeding the recovery robot from completing the task, the trial is discarded, and a new target position is subsequently sampled. This ensures the integrity of the sampling process.

\subsection{Optimization}
Based on the analyzed dataset, our proposed framework addresses the challenge of identifying optimal thresholds within the priority function. Initially, we derive regressions for the sampled data obtained from the developed HRC simulator. In this investigation, we utilize Gaussian process regression (GPR) to represent a nonlinear regression function $\mathcal{GP}$. This approach enables the estimation of productivity and time risk, along with their associated confidence, under various threshold conditions. The regression functions are expressed as $f_{product}\sim \mathcal{GP}(X_{product}) = \mathcal{N}( \mu_{product}, \sigma_{product})$ and $f_{risk}\sim \mathcal{GP}(X_{risk}) = \mathcal{N}( \mu_{risk}, \sigma_{risk})$, respectively.
The regressions are implemented using the open-source library \textit{GPy} \cite{GPy}. The kernel functions employed in this study are the Matern32 function in Section \ref{sec:hardware_exp} and the exponential function in Section \ref{sec:simulation_exp}. These choices are informed by a pilot investigation assessing regression accuracy within each experimental setup.
Subsequently, we formulate the following optimization problem to ascertain the optimal threshold $\bar{\bm{l}}^*$:
\begin{align}
    \begin{aligned}
         &               & \bar{\bm{l}}^* = & \mathop{\arg \max}_{\bar{\bm{l}}} f_{product}(\bar{\bm{l}}) \\
         & \mathop{s.t.} &                  & \ f_{risk}(\bar{\bm{l}}) < t_{lim}                          \\
         &               &                  & 0 \le \bar{l}_{i_m} \le l_{max} \ (i_m = 1 \cdots N_m)          \\
    \end{aligned}
    \label{eq:threshold_opt}
\end{align}
where $t_{lim}$ is the risk time limit provided as prior human knowledge.
The objective function seeks to maximize the estimated mean of the productivity regression, expressed as $f_{product} = \mu_{product}$. Conversely, the inequality constraint equation, which restricts the maximum risk time, adopts a conservative approach by accounting for the disparity between the simulator and real environments. This consideration is grounded in the estimated distribution of the regression function, as follows: 
\begin{equation}
    f_{risk} = \mu_{risk} + \zeta \sigma_{risk}
    \label{eq:f_risk}
\end{equation}
for practical usage.
The parameter $\zeta \ge 0$ represents a gain that quantifies the degree to which the optimization accounts for regression confidence. Presently, the determination of the confidence gain $\zeta$ relies solely on empirical trials assessing the accuracy of the productivity and risk time regressions. Subsequent research endeavors will explore a more structured methodology for identifying the optimal confidence gain $\zeta$.
To address this optimization problem, we employ a genetic algorithm (GA) implemented within the open-source library \textit{DEAP} \cite{Gagne2012}. This approach treats the problem as a black-box optimization task.

\section{Preliminary Qualitative Experiment}\label{sec:hardware_exp}

This section introduces a preliminary experiment aimed at qualitatively demonstrating the viability of the proposed framework using both an automated robot and a teleoperated robot. The experiment encompasses two primary assessments: 1) Optimization Assessment, which explores the feasibility of the presented optimization problem, and 2) Interaction Assessment, which evaluates the practicality of the proposed collision avoidance algorithm within a robot hardware system.
It is important to note that this preliminary experiment does not delve into the examination of psychological and productive effects on the human operator. Quantitative assessments of productivity and risk time are conducted in this study, leveraging the developed HRC simulator as detailed in Section \ref{sec:simulation_exp}. Subsequent investigations will focus on understanding how our proposed methodology influences human operators in psychological and productive dimensions through human experiments, subject to ethical committee approval.

\subsection{Teleoperation System}\label{sec:teleoperation_system}

\begin{figure}[t]
    \centering
    \includegraphics[width=0.9\linewidth]{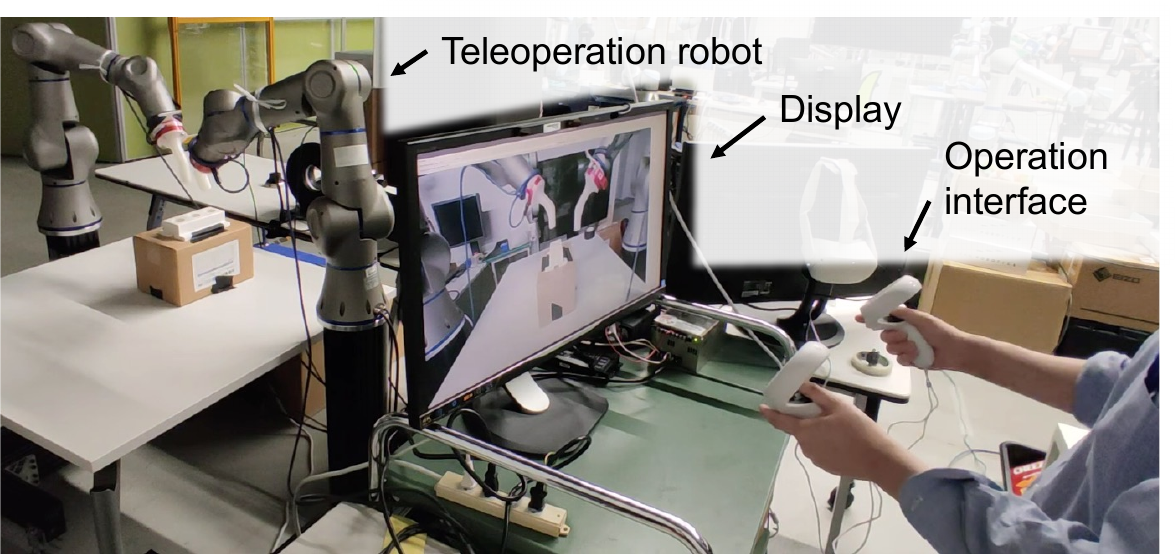}
    \caption{Snapshots of the hardware experiment environments. }
    \label{fig:hardware_exp}
\end{figure}

\begin{figure}[!t]
    \centering
    \includegraphics[width=0.99\linewidth]{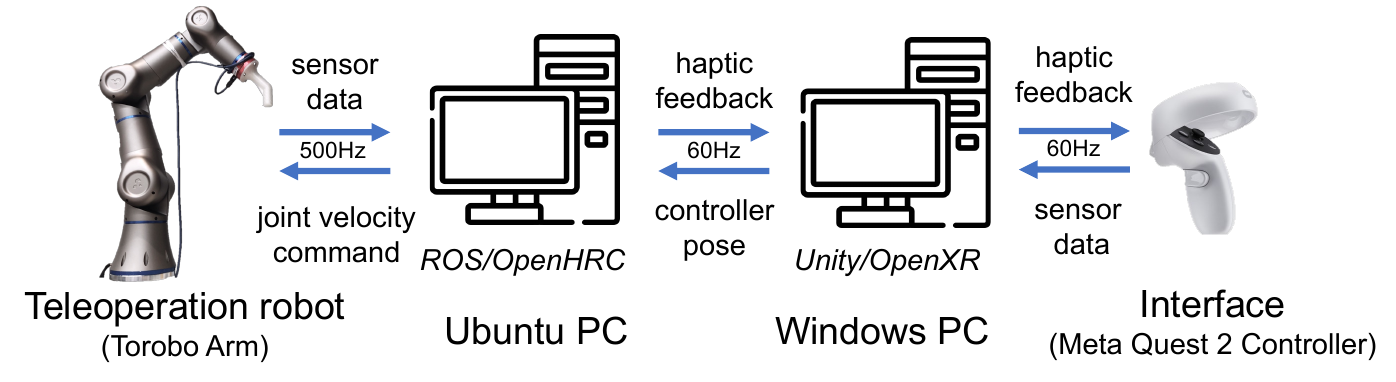}
    \caption{Teleoperation system architecture used in this study. The system is composed of a Ubuntu PC for robot control based on \textit{OpenHRC} and a Windows PC for VR controller based on \textit{Unity}.}
    \label{fig:teleoperation_system}
\end{figure}

The system utilized in this assessment is depicted in Fig.\ref{fig:hardware_exp}. For this initial evaluation, we establish a teleoperation environment, comprising a VR controller (Meta Quest 2 Touch Controller, Meta Platforms, Inc.) connected to the teleoperated robot (ToroboArm) via ROS-based communication. The architecture of our teleoperation system is outlined in Fig.\ref{fig:teleoperation_system}. The incorporation of VR devices in teleoperation systems is common, as they provide operators with a heightened sense of presence, and their integration is seamless through the use of game engines such as Unity \cite{Coronado2023}.
During teleoperation, activated by pressing the grasp button, the robot's end-effector aligns with the controller's relative position and absolute orientation, akin to the VR-based mobile manipulator teleoperation system presented in \cite{Nakanishi2020}.

\subsection{Experimental Conditions}

\begin{table}[t]
    \centering
    \caption{Parameter Configuration in Preliminary Experiment}
    \label{tab:exp0_param}
    \begin{tabular}{clr} \hline \hline
        Param    & Definition                             & Value               \\\hline
        $\xi$    & Gain for convergence speed             & 0.1                 \\
        $d_s$    & Security distance                      & 0.10                \\
        $d_i$    & Influenced distance                    & 0.15                \\\hline
        $M$      & Virtual mass                           & diag(1, 1, 1)       \\
        $K$    & Virtual spring for recovery            & diag(40, 40, 20)    \\\hline
        $w_o$    & Nominal weight for IK                  & $1.0 \times 10^5 $  \\
        $\gamma$ & Gain for prioritization                & $1.0 \times 10^2 $  \\\hline \hline
    \end{tabular}
\end{table}

The preliminary experiment employs two robots, namely a standalone manufacturing robot and a teleoperated recovery robot, to showcase the viability of the proposed framework within a streamlined configuration, which corresponds to $N_r = 1$ and $N_m = 1$ in Section \ref{sec:Implementation}. 
To simplify the notation, this section omits the robot index, e.g., the relative distance for the priority distance $l_{1}$ and its threshold $\bar{l}_{1}$ are denoted as $l$ and $\bar{l}$, respectively.

For ease of explanation, we separate the qualitative assessment of the proposed framework into two sub-assessments in the following sections.

\subsubsection{Optimization Assessment}
The first sub-assessment examines the validity of our proposed optimization framework. The configuration parameters for the robot controllers and data sampling are detailed in Table \ref{tab:exp0_param}. During the data collection phase, a dataset comprises the mean of 100 reaching trials executed by the recovery robot, employing a randomly selected threshold. Subsequently, we gather 50 samples from the HRC simulator, each associated with distinct thresholds.
In the optimization phase, we designate regression confidence gains as $\zeta = 0, 1, 2, 3$ in Eq.~(\ref{eq:f_risk}). For each confidence gain value, the optimization procedure is executed across multiple risk time limits $t_{lim} \in [2.3,\ 2.7]$, segmented at 0.01 s intervals. Our evaluation encompasses the sampled productivity and risk time data, their respective regressions, and the resultant optimization results.

\subsubsection{Interaction Assessment}
\begin{figure}[!t]
    \centering
    \includegraphics[width=0.7\linewidth]{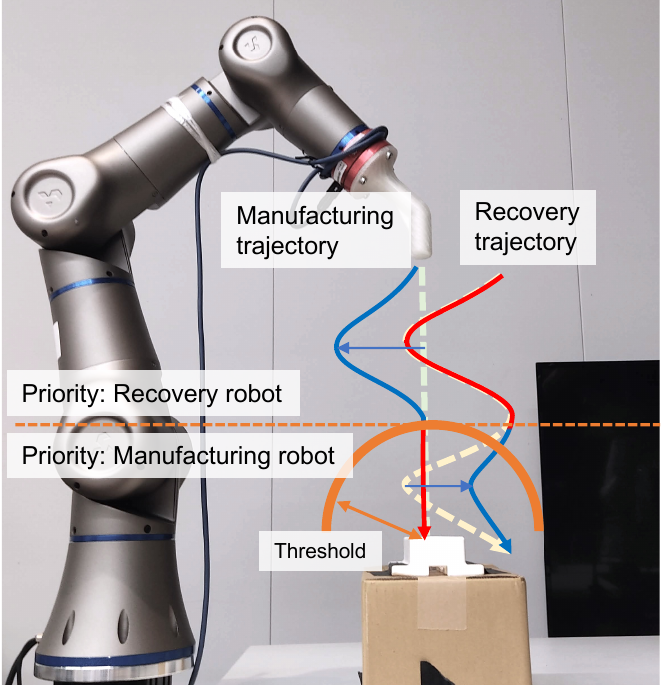}
    \caption{Target/expected and actual trajectories of the autonomous manufacturing and teleoperated recovery robots in the preliminary interaction assessment, represented by solid and dashed lines, respectively. The manufacturing robot is programmed to descend vertically, while the recovery robot is directed to initiate two collisions with the manufacturing counterpart. The red parts of the trajectories indicate that the corresponding robot is prioritized over the other. An orange semicircle demarcates the threshold at which priority switching occurs.}
    \label{fig:exp0_trajectory_target}
\end{figure}

In the second sub-assessment, we investigate the feasibility of the proposed control framework with a hardware system.
As a preliminary assumption, we consider the priority threshold to be $\bar{l} = 0.15$\, and the initial position of the manufacturing robot's end-effector is set at 0.3\,m above its target.
During the assessment, we guide the teleoperated robot's end-effector to collide twice with the manufacturing robot's end-effector, once above and once below the specified threshold height, as illustrated in Fig.~\ref{fig:exp0_trajectory_target}. The evaluation of feasibility is conducted qualitatively, based on the observed trajectories of the robots.

\subsection{Experimental Results}
\subsubsection{Optimization Assessment}

\begin{figure}[t]
    \centering
    \begin{minipage}[b]{0.49\hsize}
        \centering
        \includegraphics[clip,width=0.99\linewidth]{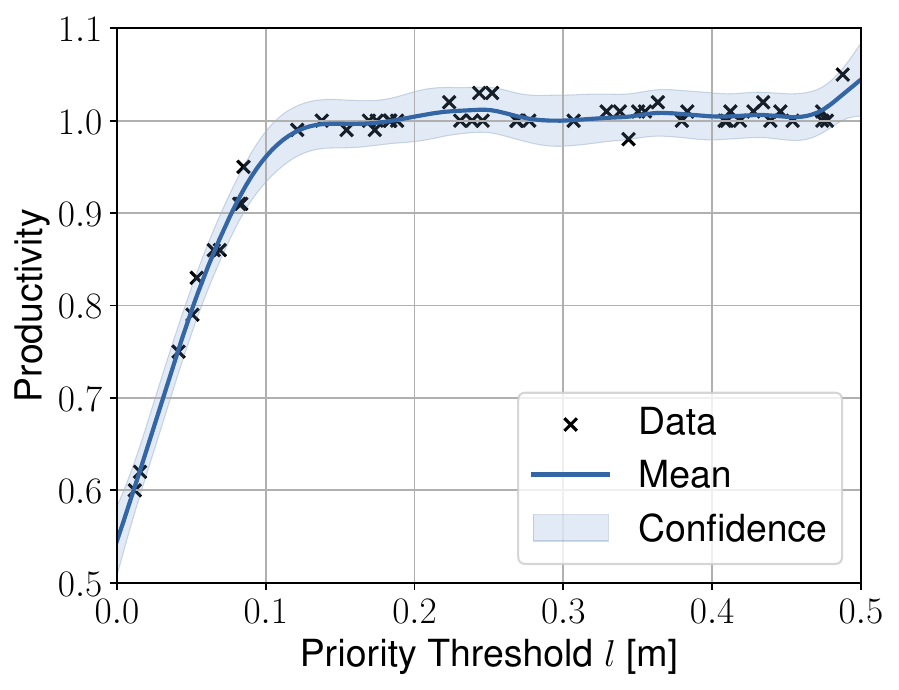}
        \subcaption{Sampled productivity data $X_{product}$.}
    \end{minipage}
    \begin{minipage}[b]{0.49\hsize}
        \centering
        \includegraphics[clip,width=0.99\linewidth]{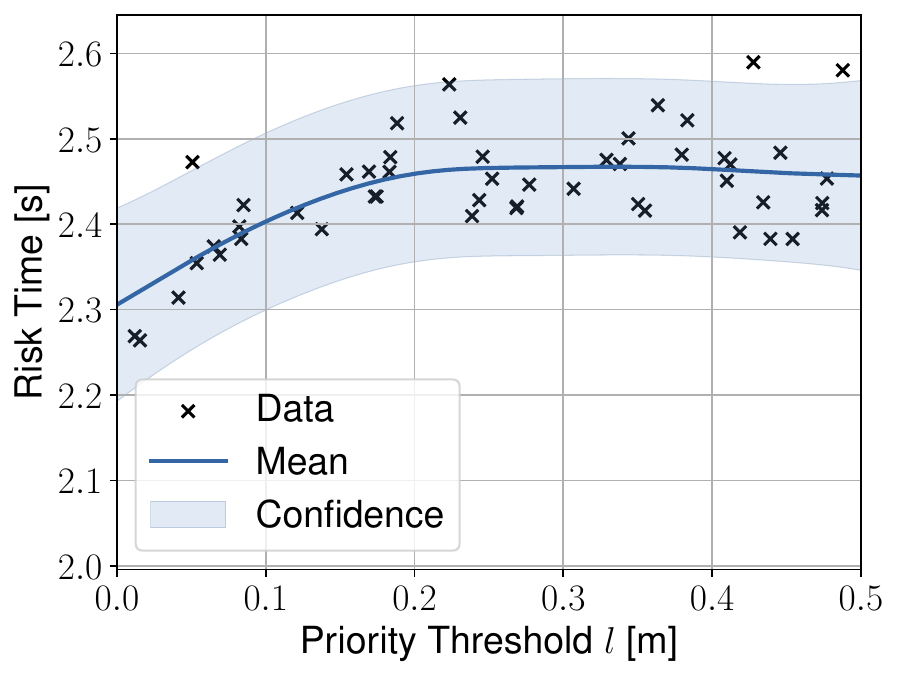}
        \subcaption{Sampled risk time data $X_{risk}$.}
    \end{minipage}
    \caption{Sampled (a) productivity data $X_{product}$ and (b) risk time data $X_{risk}$ from the HRC simulation composed of one each of manufacturing and recovery robots in the preliminary optimization assessment. In addition, the blue lines and areas illustrate the means and distributions of the obtained GPR functions, respectively. The horizontal axis represents the threshold for switching the prioritized robot.}
    \label{fig:exp0_data}
\end{figure}

\begin{figure}[!t]
    \centering
    \includegraphics[width=0.8\linewidth]{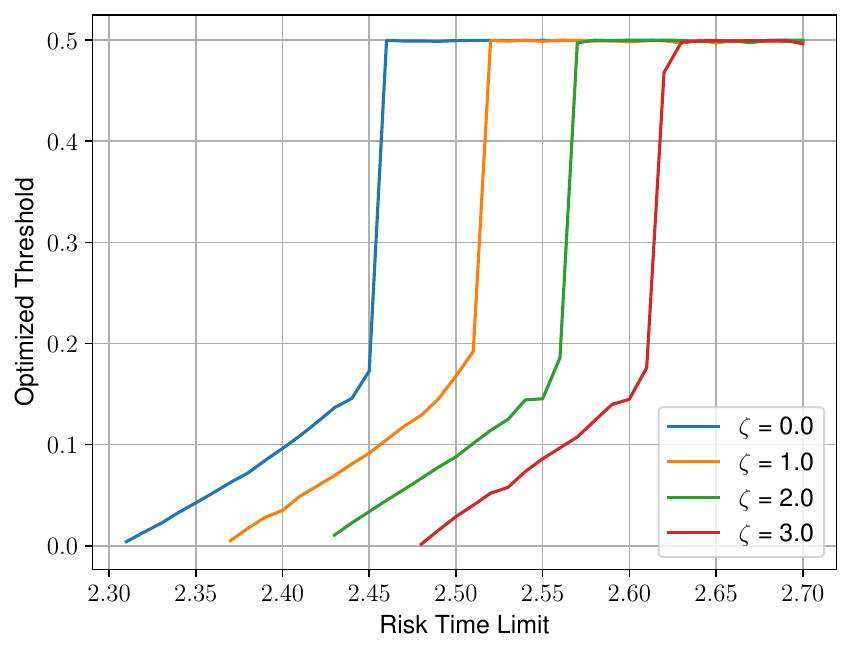}
    \caption{Optimized priority threshold $\bar{l}^*$ in the preliminary optimization assessment. The risk time limits are chosen from $[2.3,\ 2.7]$ at intervals of 0.01\,s. The line color indicates the regression confidence gain $\zeta$. A higher regression confidence gain $\zeta$ corresponds to a more extensive range of thresholds identified by the optimizer.}
    \label{fig:exp0_opt}
\end{figure}

Figure \ref{fig:exp0_data} presents the collected data samples alongside the Gaussian regression functions for risk time and productivity. A discernible trend emerges, indicating that both productivity and risk time exhibit an increase with larger priority thresholds. This trend aligns with our research motivation, affirming that our priority optimization can effectively govern the trade-off between productivity and risk time. Notably, the data points corresponding to $\bar{l} = 0$ reflect scenarios where the recovery robot is consistently prioritized, while saturation near $\bar{l} = l_{max}$ suggests situations where the manufacturing robot holds constant priority.
This observation suggests the potential identification of an optimal priority threshold $\bar{l}^*$ that maximizes productivity while adhering to the risk time limit during the transition phase $\bar{l} \in [0,\ 1]$.
Figure \ref{fig:exp0_opt} illustrates the optimized priority threshold, with the horizontal axis representing the risk time limit $t_{lim}$ in eq.~(\ref{eq:threshold_opt}). As intended, a higher confidence gain $\zeta$ results in a broader range of thresholds identified by the optimizer. While data trends may become more intricate in systems featuring multiple manufacturing robots, our approach exhibits the potential to capture the nuanced features of productivity and risk time. This suggests its applicability in practical scenarios involving complex systems.

\subsubsection{Interaction Assessment}

\begin{figure}[!t]
    \centering
    \includegraphics[width=0.95\linewidth]{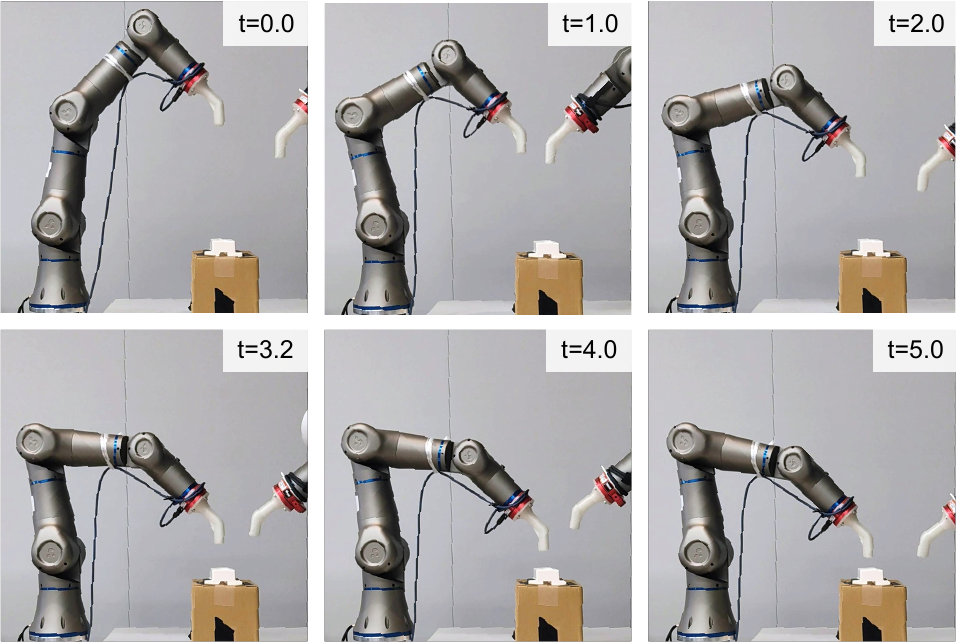}
    \caption{Snapshots in the preliminary interaction assessment.}
    \label{fig:exp0_shapshot}
\end{figure}
\begin{figure}[!t]
    \centering
    \includegraphics[width=0.95\linewidth]{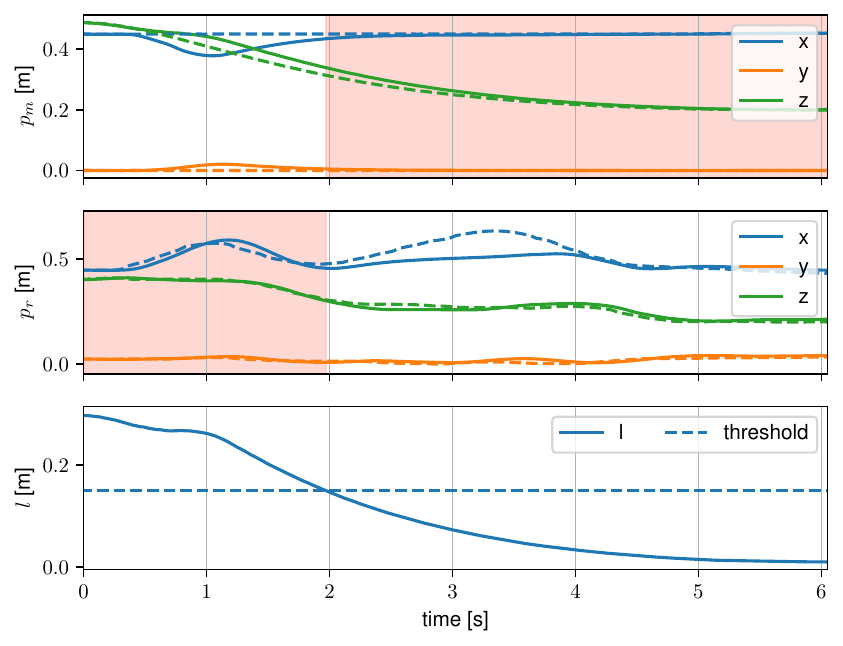}
    \caption{Resultant trajectories of the autonomous manufacturing robot $\bm{p}_m$ (first row) and the teleoperated recovery robot $\bm{p}_r$ (second row) in the preliminary interaction assessment. The solid and dashed lines indicate the actual and target/expected trajectories, respectively. The orange background indicates the time duration over which the corresponding robot is prioritized. In the third graph, the solid line indicates the relative distance $l$ between the manufacturing robot end-effector and its target, while the dashed line indicates its threshold $\bar{l}$.}
    \label{fig:exp0_trajectories}
\end{figure}

Figure \ref{fig:exp0_shapshot} provides visual snapshots from the preliminary interaction assessment. In Figure \ref{fig:exp0_trajectories}, both the actual trajectories, depicted as solid lines, and the target trajectories (i.e., those expected in the absence of interaction), represented by dashed lines, for both the manufacturing and recovery robots are displayed. Additionally, the relative distance between the manufacturing robot's end-effector (solid line) and its threshold (dashed line) is illustrated.
Notably, at approximately $t=1.0$ and $t=3.2$, the operator guided the recovery robot to engage with the manufacturing robot. When the relative distance exceeded the threshold around $t=1.0$, priority was assigned to the recovery robot. Consequently, the trajectory of the manufacturing robot was adjusted to circumvent potential collisions. Conversely, around $t=3.2$, when the relative distance fell below its threshold value, the manufacturing robot was accorded priority, leading to a temporary suspension of teleoperation commands to the recovery robot to prevent collisions. In both instances, the robots effectively reverted to tracking their target positions once the relative distance expanded sufficiently.

Based on these outcomes, our proposed framework adeptly prioritizes robot actions in alignment with the relative distance parameter embedded within the priority function.
Given the successful outcomes of the preliminary experiment, which was conducted using a simplified configuration involving two robots, we proceed to a more comprehensive simulation-based experiment involving four robots, as detailed in Section \ref{sec:simulation_exp}.

\section{Quantitative Experiment}\label{sec:simulation_exp}
This section introduces experimental assessments of the proposed framework involving four robots. The aim is to quantitatively demonstrate the proof of concept for productivity-based motion priority optimization. The evaluation of optimization quality is conducted utilizing two datasets of human reaching motion, which are generated through the developed HRC simulator.

\subsection{Experimental Conditions}
\begin{table}[tb]
    \centering
    \caption{Parameter Configuration in Experiments}
    \label{tab:param}
    \begin{tabular}{clr} \hline \hline
        Param    & Definition                             & Value               \\\hline
        $\xi$    & Gain for convergence speed             & 0.1                 \\
        $d_s$    & Security distance                      & 0.10                \\
        $d_i$    & Influenced distance                    & 0.15                \\\hline
        $M$      & Virtual mass                           & diag(1, 1, 1)       \\
        $K_r$    & Virtual spring for recovery            & diag(15, 15, 15)    \\
        $K_{m1}$ & Virtual spring for manufacturing (\#1) & diag(30, 30, 20) \\
        $K_{m2}$ & Virtual spring for manufacturing (\#2) & diag(40, 40, 30)     \\
        $K_{m3}$ & Virtual spring for manufacturing (\#3) & diag(50, 50, 40)    \\\hline
        $w_o$    & Nominal weight for IK                  & $1.0 \times 10^4 $  \\
        $\gamma$ & Gain for prioritization                & $1.0 \times 10^2 $  \\
        $\zeta$  & Gain for regression confidence         & 0 \\\hline \hline
    \end{tabular}
\end{table}

As illustrated in Fig.\,\ref{fig:recovery_situation}, the developed simulator encompasses one teleoperated recovery robot ($N_r = 1$) and three autonomous manufacturing robots ($N_m=3$). We leverage two ProMPs models, namely model-A and model-B, trained with different trajectory types as discussed in Section \ref{sec:promp}. In this context, we refer to the simulators using model-A and model-B as sim-A and sim-B, respectively.
The parameter configurations for the robot controllers and data sampling are outlined in Table \ref{tab:param}. Notably, the regression confidence gain is set to zero ($\zeta = 0$) for this assessment, emphasizing the focus on framework feasibility rather than practical performance.
Similar to the optimization assessment in Section \ref{sec:hardware_exp}, each data sample corresponds to the mean of 100 reaching trials executed by the recovery robot. To determine a suitable range for the risk time limit, three preliminary benchmark tests are conducted, collecting 100 data samples each in the following scenarios: 1) Non-continuous (NC) case, where recovery and manufacturing processes are separated; 2) Always prioritizing recovery (PR) case; and 3) Always prioritizing manufacturing (PM) case. The NC case corresponds to conventional manufacturing, as depicted in Fig.\,\ref{fig:background}, where the manufacturing process halts until the human operator completes recovery. The PR and PM conditions represent scenarios with $\bar{\bm{l}}^* = \bm{0}$ and $\bar{\bm{l}}^* = \bm{1} \cdot \bar{l}_{max}$, respectively.
Furthermore, we collect 500 samples from the HRC simulator with randomly chosen thresholds for the proposed optimization in eq.\,(\ref{eq:threshold_opt}). The optimization is executed using the accumulated 700 data samples, encompassing the random threshold case (500 samples), PM case (100 samples), and PR case (100 samples).
To assess the optimization results, 100 samples are collected from the proposed simulator with the optimized thresholds $\bar{\bm{l}}^*$, and the productivity and risk time are evaluated to confirm the efficacy of the proposed simulation-based motion priority optimization.

\subsection{Experimental Results}

\begin{figure}[t]
    \centering
    \begin{minipage}[b]{0.99\hsize}
        \centering
        \includegraphics[clip,width=0.99\linewidth]{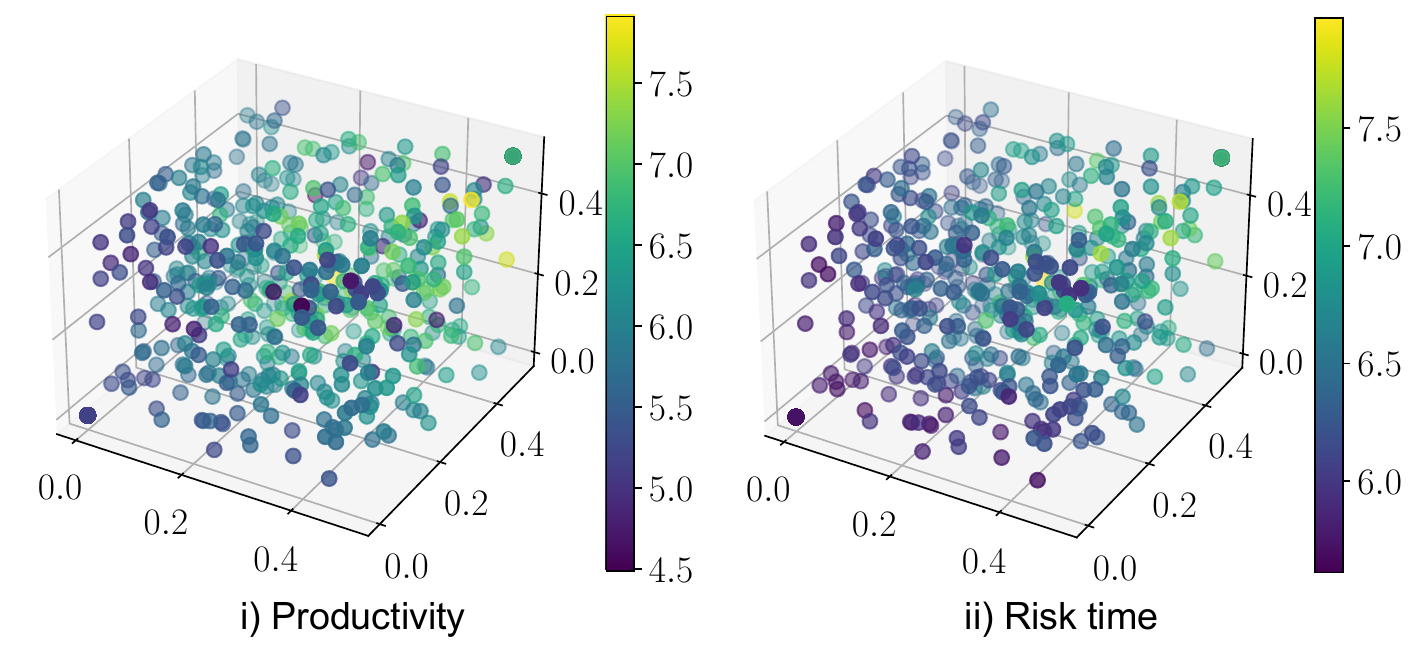}
        \subcaption{Dataset collected from sim-A.}
    \end{minipage}\\\vspace{3mm}
    \begin{minipage}[b]{0.99\hsize}
        \centering
        \includegraphics[clip,width=0.99\linewidth]{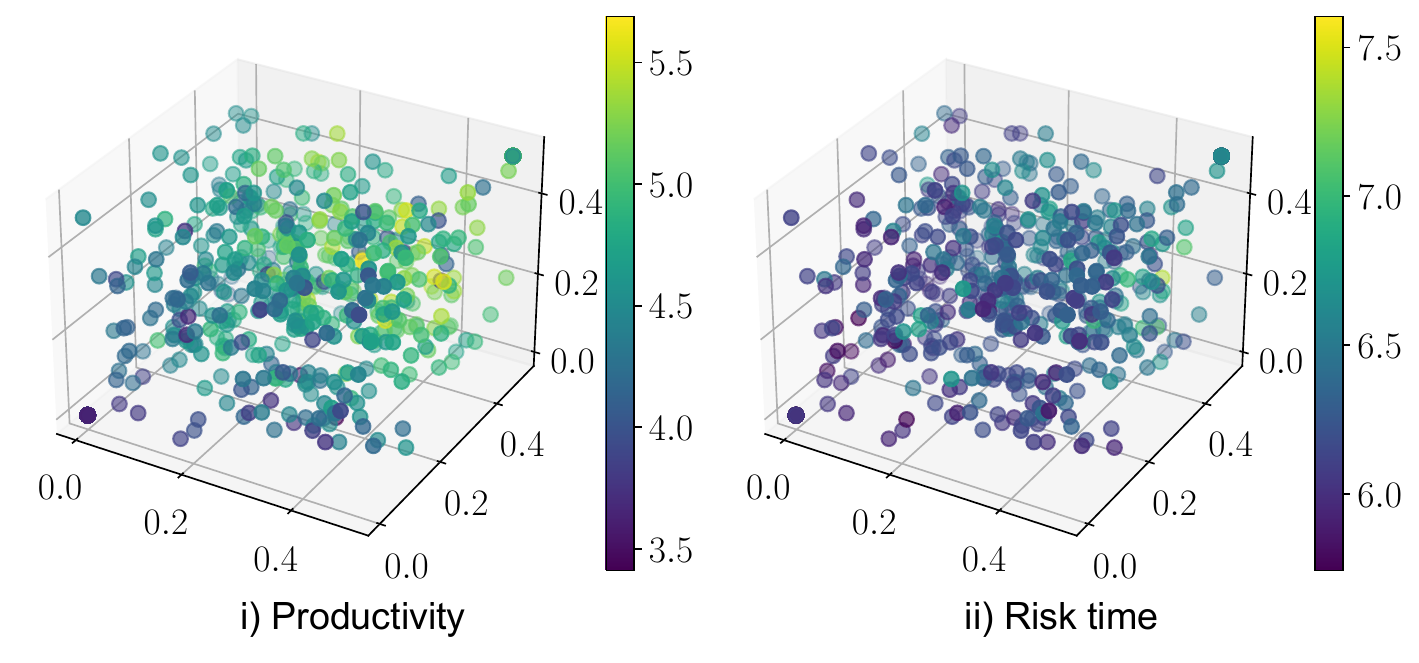}
        \subcaption{Dataset collected from sim-B.}
    \end{minipage}
    \caption{Sampled productivity data $X_{product}$ (left figures) and risk time data $X_{risk}$ (right figures) from the HRC simulations; (a) sim-A and (b) sim-B. Each axis corresponds to the thresholds $\bar{l}_{i_m}$ ($i_m = 1 \cdots 3$) for the priority function in eq.\,(\ref{eq:priotiry}). The colors of the sampled data represent the productivity and risk time values under each threshold condition.}
    \label{fig:data}
\end{figure}

\begin{table}[t]
    \centering
    \caption{Preliminary Benchmark Results (Mean $\pm$ SD)}
    \label{tab:benchmark}
    \begin{tabular}{lcc} \hline \hline
        & \multicolumn{2}{c}{sim-A} \\
        Conditions                        & Productivity    & Risk Time       \\\hline
        Non-continuous (NC)                   & -               & 5.57 $\pm$ 0.06 \\
        Prioritizing Recovery (PR)     & 5.17 $\pm$ 0.08 & 5.75 $\pm$ 0.08 \\
        Prioritizing Manufacturing (PM) & 6.45 $\pm$ 0.55 & 6.88 $\pm$ 0.32 \\\hline \hline
        & \multicolumn{2}{c}{sim-B} \\
        Conditions                        & Productivity    & Risk Time       \\\hline
        Non-continuous (NC)                   & -               & 5.75 $\pm$ 0.07 \\
        Prioritizing Recovery (PR)     & 3.74 $\pm$ 0.12 & 5.97 $\pm$ 0.10 \\
        Prioritizing Manufacturing (PM) & 4.89 $\pm$ 0.34 & 6.55 $\pm$ 0.25 \\\hline \hline
    \end{tabular}
\end{table}

\begin{table*}[t]
    \centering
    \caption{Optimization Results of Thresholds $\bar{\bm{l}}^*$}
    \label{tab:results}
    \begin{tabular}{c|c|c|c|c|c|c|c|c|c|c|c|c|c|c|c|c|c|c} \hline \hline
            opt-A          & \multicolumn{18}{c}{Risk Time Limit $t_{lim}$ [s]}          \\
                       & 5.8  & 5.9  & 6.0  & 6.1  & 6.2  & 6.3 & 6.4  & 6.5  & 6.6  & 6.7  & 6.8  & 6.9  & 7.0  & 7.1  & 7.2 & 7.3 &7.4 & 7.5       \\\hline
        $\bar{l}_1^*$    & 0.03 &0.05  &0.08  &0.09  &0.05  &0.07  &0.13  &0.11  &0.16  &0.18  &0.21  &0.33 &0.26  &0.28  &0.28&0.28&0.28 &0.28\\
        $\bar{l}_2^*$    &0.06&0.10&0.12&0.14&0.34&0.32&0.19&0.30&0.24&0.26&0.26&0.40&0.32&0.34&0.34&0.34&0.34 &0.34 \\
        $\bar{l}_3^*$    &0.11&0.14&0.15&0.14&0.12&0.14&0.15&0.15&0.15&0.15&0.15&0.13&0.15&0.16&0.16&0.16&0.16 &  0.16\\\hline \hline
        opt-B          & \multicolumn{18}{c}{Risk Time Limit $t_{lim}$ [s]}          \\
                       & 5.8  & 5.9  & 6.0  & 6.1  & 6.2  & 6.3 & 6.4  & 6.5  & 6.6  & 6.7  & 6.8  & 6.9  & 7.0  & 7.1  & 7.2 & 7.3 &7.4   & 7.5       \\\hline
        $\bar{l}_1^*$   & -    & -    & 0.0&0.02&0.08&0.43&0.47&0.46&0.43&0.43&0.43&0.43&0.43&0.43&0.43&0.43&0.43&0.43 \\
        $\bar{l}_2^*$    & -    & -    & 0.11&0.3&0.45&0.2&0.28&0.38&0.45&0.45&0.45&0.45&0.45&0.45&0.45&0.45&0.45&0.45 \\
        $\bar{l}_3^*$   & -    & -    & 0.06&0.17&0.38&0.34&0.32&0.16&0.17&0.17&0.17&0.17&0.17&0.17&0.17&0.17&0.17&0.17 \\\hline \hline
    \end{tabular}
\end{table*}

\begin{figure*}[t]
    \centering
    \begin{minipage}[b]{0.495\hsize}
        \centering
        \includegraphics[clip,width=0.99\linewidth]{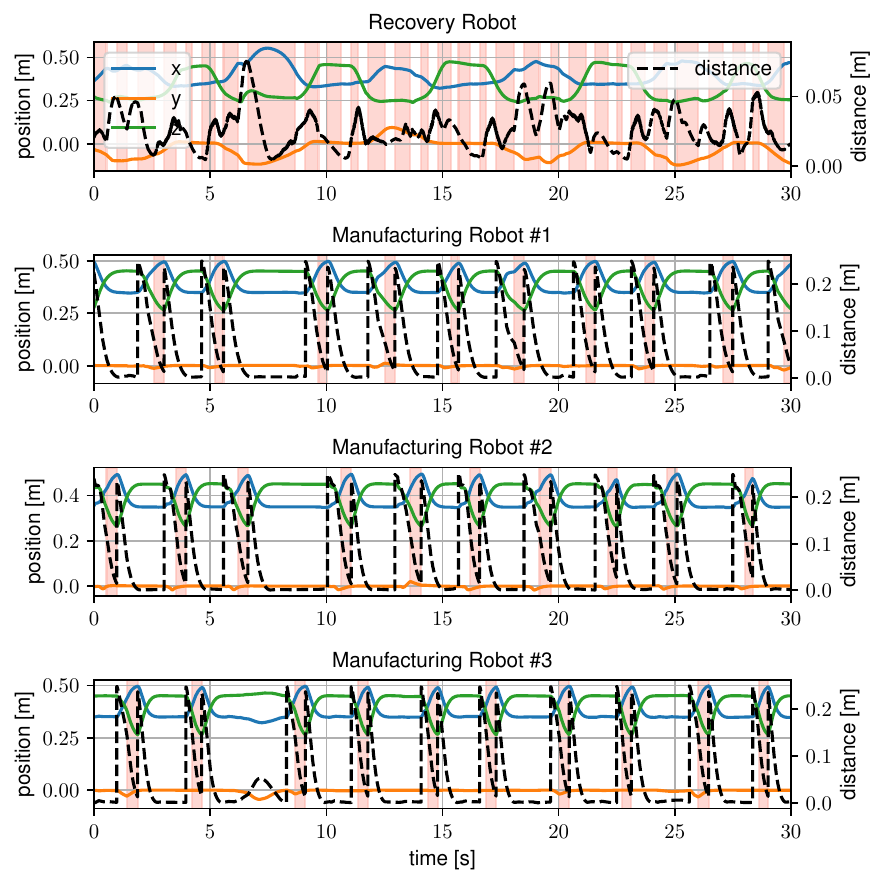}
        \subcaption{Assessment in sim-A using the parameters optimized by opt-A with $t_{lim} = 6.0$.}
    \end{minipage}
    \begin{minipage}[b]{0.495\hsize}
        \centering
        \includegraphics[clip,width=0.99\linewidth]{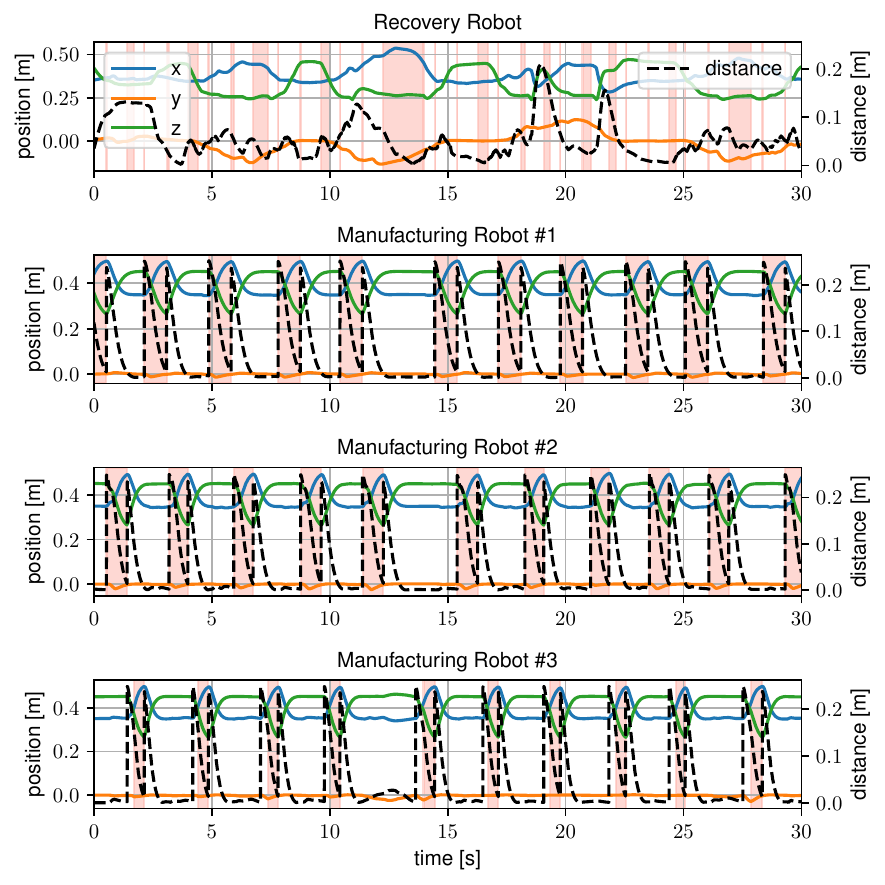}
        \subcaption{Assessment in sim-A using the parameters optimized by opt-A with $t_{lim} = 7.5$.}
    \end{minipage}\\ \vspace{3mm}
    \begin{minipage}[b]{0.495\hsize}
        \centering
        \includegraphics[clip,width=0.99\linewidth]{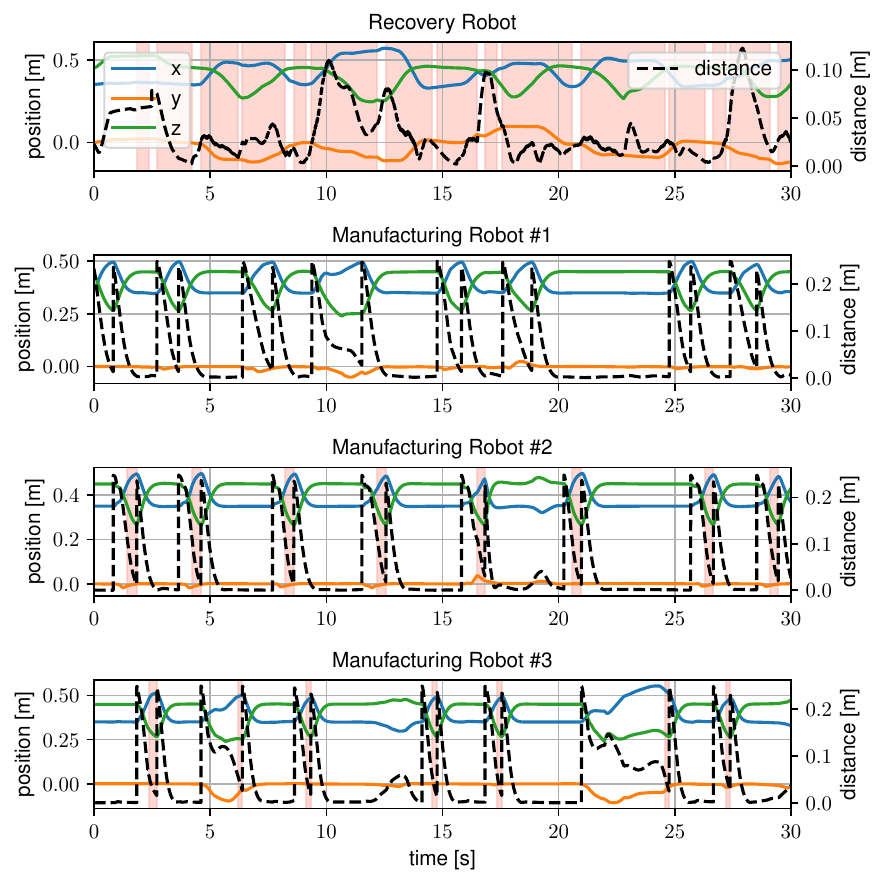}
        \subcaption{Assessment in sim-B using the parameters optimized by opt-B with $t_{lim} = 6.0$.}
    \end{minipage}
    \begin{minipage}[b]{0.495\hsize}
        \centering
        \includegraphics[clip,width=0.99\linewidth]{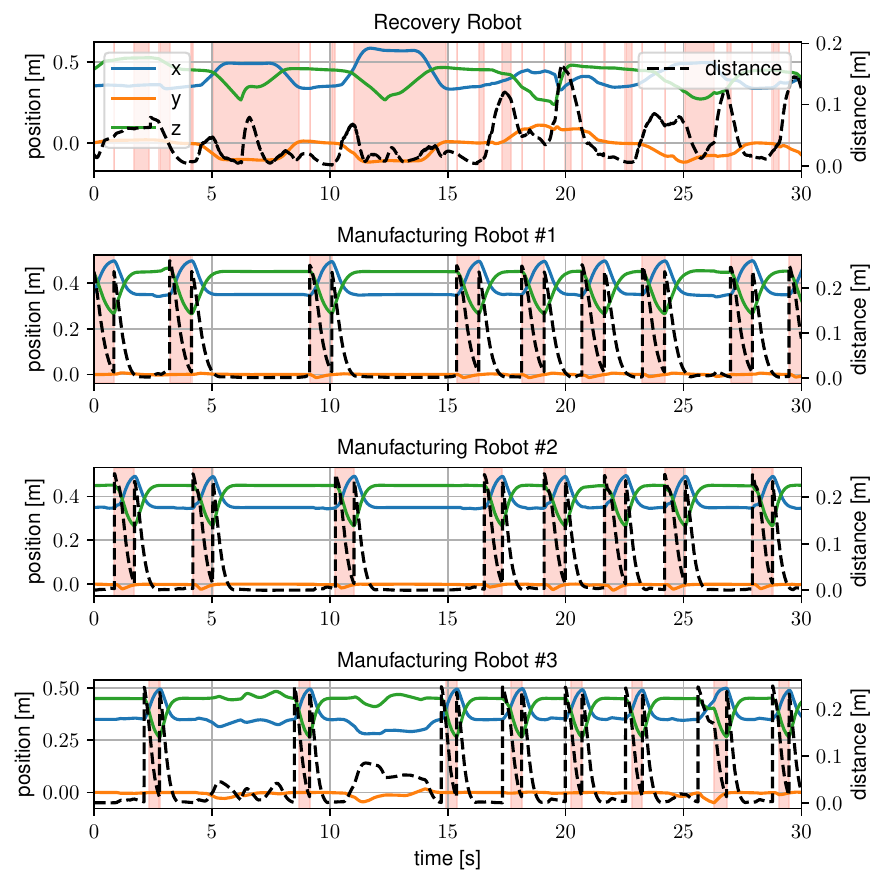}
        \subcaption{Assessment in sim-B using the parameters optimized by opt-B with $t_{lim} = 7.5$.}
    \end{minipage}
    \caption{Robot trajectories in the HRC simulators, with solid lines representing the actual robot trajectories and dashed lines indicating the relative distances to their respective targets. Periods where the robot is prioritized are highlighted with an orange background. For consistent comparison, the same random seeds are employed for both recovery targets and motion generators across all conditions. Priority function thresholds are determined based on risk time limits of 6.0 and 7.5, as elaborated in Table \ref{tab:results}.}
    \label{fig:trajectory}
\end{figure*}

\begin{figure}[t]
    \centering
    \begin{minipage}[b]{0.99\hsize}
        \centering
        \includegraphics[clip,width=0.99\linewidth]{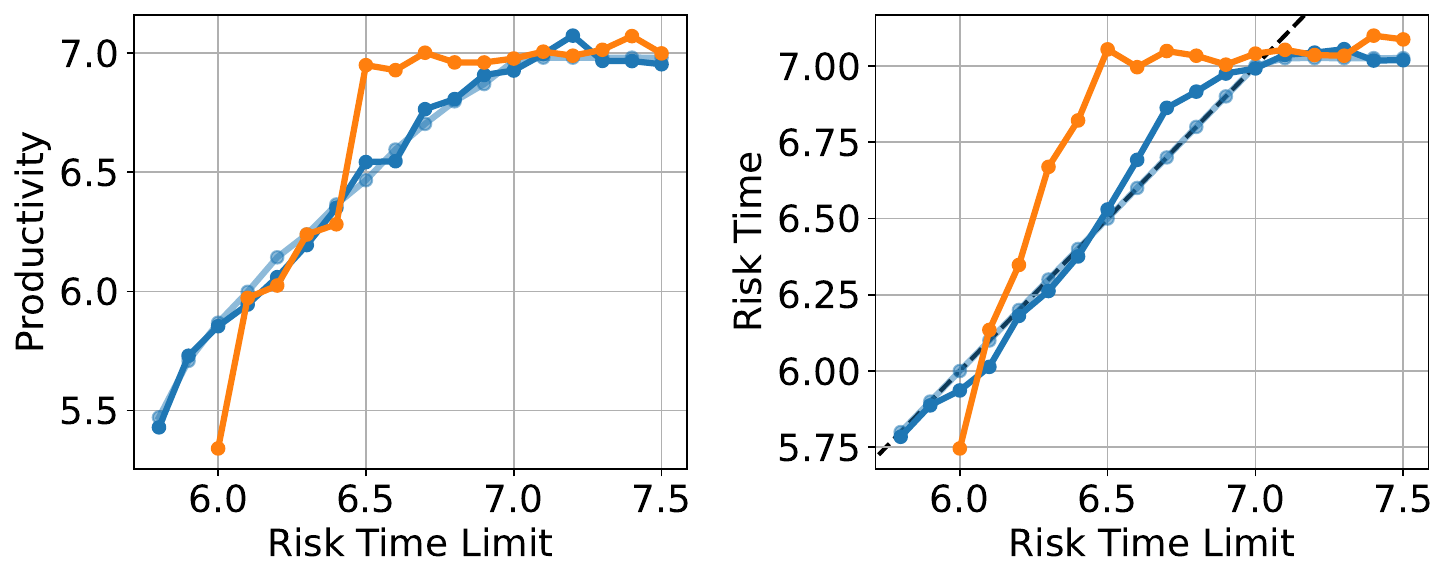}
        \subcaption{Assessment of the optimized parameters in sim-A.}
        \label{fig:assessment_A}
    \end{minipage}\\ \vspace{3mm}
    \begin{minipage}[b]{0.99\hsize}
        \centering
        \includegraphics[clip,width=0.99\linewidth]{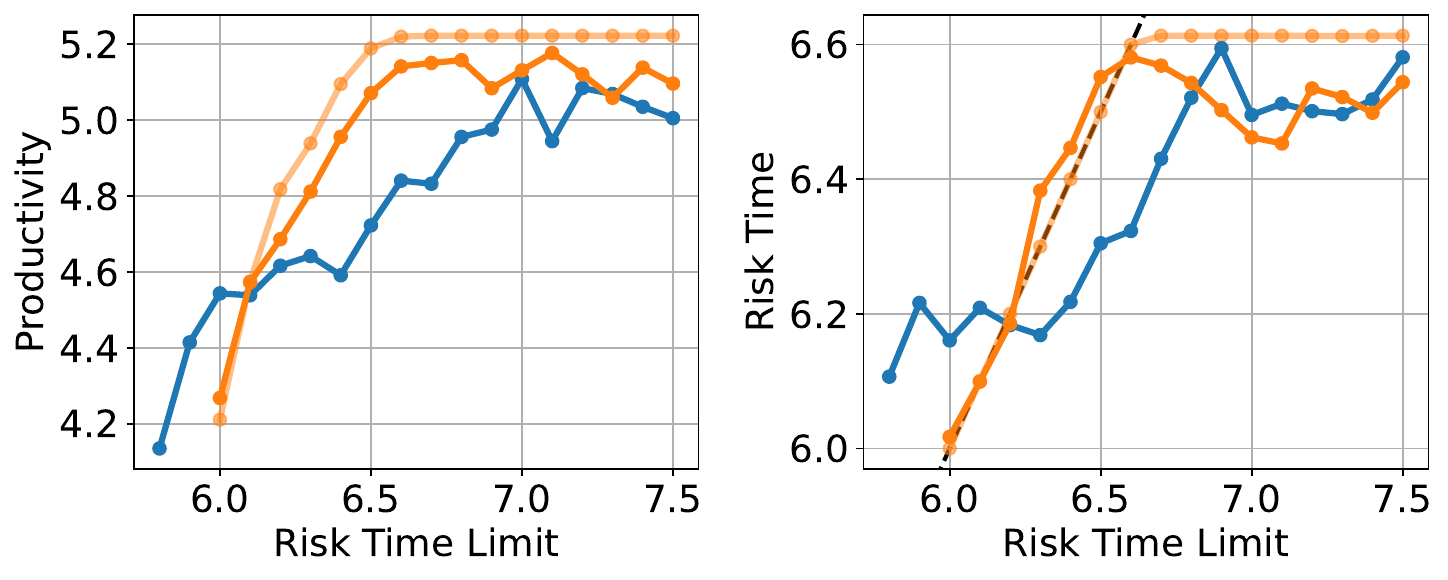}
        \subcaption{Assessment of the optimized parameters in sim-B.}
        \label{fig:assessment_B}
    \end{minipage}
    \caption{ Means of productivity and risk time within the designated risk time limits utilizing the corresponding optimized parameters are presented. The light-colored lines denote the output of the optimization process. Productivity and risk time evaluations are conducted in (a) sim-A and (b) sim-B using the two optimized parameter sets from opt-A and opt-B. Notably, results originating from the same source, i.e., blue lines in (a) and orange lines in (b), exhibit relatively close values. Conversely, outcomes under opposite source conditions, i.e., the orange line in (a) and blue lines in (b), indicate lower productivity and surpass the risk time limit. These findings suggest that the proposed method effectively captures distinctive features of the recovery motion generators and successfully identifies optimal motion priority functions for each human demonstration.}
    \label{fig:assessment}
\end{figure}

Figure \ref{fig:data} illustrates the collected data on productivity $X_{product}$ and risk time $X_{risk}$ obtained using sim-A and sim-B. The results of the preliminary benchmark tests are presented in Table \ref{tab:benchmark}.
In the NC case, where the recovery robot does not interfere with the manufacturing process, the risk time is theoretically minimal, but productivity is zero. By transitioning to continuous recovery processes and consistently prioritizing the recovery robot (PR case), productivity improved, accompanied by a 3.2\% increase in risk time in sim-A and a 3.8\% increase in sim-B.
Comparatively, the PM case, where manufacturing is always prioritized, resulted in a 24.8\% improvement in productivity in sim-A and a 30.7\% improvement in sim-B compared to the PR case. However, this came at the cost of a 19.7\% increase in risk time in sim-A and a 9.7\% increase in sim-B.
These outcomes highlight the non-continuous trade-off between productivity and risk time when introducing teleoperation to recovery scenarios, as illustrated in the left and middle columns of Fig.\,\ref{fig:background}. Given our objective of finding optimal thresholds interpolating PR and PM cases, we selected the test set of risk time limits as $t_{lim} = [5.8, 7.5]$ at intervals of 0.1\,s.


Table \ref{tab:results} displays the optimization results of the priority function thresholds $\bar{\bm{l}}^*$ for the selected risk time limits $t_{lim}$. A dash mark (-) indicates that the optimizer failed to find a feasible solution meeting the risk time limit constraint. Such unfeasible cases are reasonable, given that these risk time limits are lower than the measured risk time in the PR case. In the remaining cases, the proposed optimization successfully identified feasible solutions.
Figure \ref{fig:trajectory} illustrates robot trajectories in the HRC simulators utilizing the optimized thresholds $\bar{\bm{l}}^*$ for $t_{lim} = 6.0$ and $7.5$. When the risk time limit $t_{lim}$ is set to 6.0, priority is almost always assigned to the recovery robot, reflecting a more conservative approach to prioritizing safety. Conversely, when $t_{lim}$ is 7.5, the prioritized time shifts towards the manufacturing robots to minimize productivity loss.
In this experimental condition, 60.1\% (sim-A) and 43.2\% (sim-B) of the time duration that the manufacturing robot prioritized in the case of $t_{lim} = 7.5$ is reallocated to the recovery robot in the case of $t_{lim} = 6.0$. 
Figure \ref{fig:assessment} presents the assessment results of the optimized motion priority function in the HRC simulator. Blue and orange lines represent the results using the optimization parameters for opt-A and opt-B, respectively. Figures (a) and (b) in Fig.~\ref{fig:assessment} correspond to the evaluation in sim-A and sim-B, respectively. The light-colored lines depict the optimization outcomes of expected productivity and risk time, i.e., $f_{product}(\bar{\bm{l}}^*)$ and $f_{risk}(\bar{\bm{l}}^*)$.
The proximity of the blue lines in Fig.~\ref{fig:assessment_A} and the orange lines in \ref{fig:assessment_B} indicates that the proposed optimization method successfully identified reasonable priority thresholds. Conversely, opposite color lines, i.e., the orange line in Fig.~\ref{fig:assessment_A} and the blue line in \ref{fig:assessment_B}, show inferior performance, such as lower productivity and exceeded risk time. This outcome suggests that the optimization results adapted to each recovery motion generator, i.e., imitation models trained with different human demonstrations (model-A and model-B).

These results affirm that our proposed framework can successfully determine optimal thresholds that maximize productivity while adhering to the risk time limit, effectively interpolating between the PR and PM cases. 
The presented outcomes establish the feasibility of our proposed framework for simulation-based motion priority optimization, facilitating efficient collaboration between automated manufacturing processes and teleoperated recovery robots.
In practical terms, enhancing regression accuracy or considering the regression distribution could lead to a more comprehensive optimization solution. 
Our future work will involve applying the proposed framework to more realistic settings and exploring how it effectively balances productivity and risk time in various scenarios. 
Furthermore, we plan to make the framework compatible with online priority optimization, thereby improving computational efficiency.

\section{Conclusion}\label{sec:conclusion}
We have formulated an extensive framework to optimize motion priority between autonomous manufacturing robots and teleoperated recovery robots for achieving cooperative tele-recovery in industrial recovery settings. This framework integrates an HRC simulator, which estimates both productivity and risk time, utilizing manufacturing and recovery motion generators. Moreover, the framework employs an optimizer to ascertain the optimal motion priority function, with the goal of maximizing productivity while respecting risk time constraints. The adaptability of the motion generators for manufacturing and recovery, along with the structure of the priority function, constitute pivotal features of our framework. 
Conversely, the proposed framework operates under several assumptions, such as the availability of motion generators for manufacturing and recovery, which will be addressed in future work.
For the initial implementation of this framework, we utilized impedance control for the manufacturing motion generator and ProMPs for the recovery motion generator. The motion priority function was designed as a straightforward function.

To provide a preliminary assessment of the framework's viability, we executed an experiment involving one automated robot and one teleoperated robot. The findings corroborated that our framework efficiently prioritizes robots based on the relative proximity between the manufacturing robot and its designated target. Subsequently, a quantitative experiment with four robots substantiated the proof of concept for our framework. We assessed the optimization efficacy in the HRC simulator, utilizing two datasets of human-reaching motions. Notably, when we imposed stringent risk time constraints to prioritize safety, the framework accorded greater precedence to the recovery robot. In contrast, when aiming to minimize productivity loss with more generous risk time limits, the manufacturing robots received increased priority. These outcomes validate the framework's capability to discern optimal thresholds that harmonize productivity and risk time considerations.

This manuscript delineates a preliminary instantiation of our framework, with each constituent component amenable to adaptation for diverse methodologies, thereby facilitating applicability in real-world scenarios. Enhancing the precision of regression or integrating regression distribution could yield more holistic optimization strategies for practical deployment. Our subsequent endeavors will focus on extending the framework to intricate scenarios and assessing its efficacy therein. Real-time adjustment of priority function thresholds, potentially leveraging Gaussian optimization, represents a promising avenue for pragmatic implementation. Furthermore, we envisage conducting empirical analyses in photorealistic simulations and authentic manufacturing settings, utilizing methodologies akin to those in \cite{Tang2023}. From an ergonomics vantage point, the incorporation of pseudo-collision feedback for human operators could augment their interaction experience. Subject to ethical committee approval, we aim to examine the psychological ramifications of this system on human operators, particularly when motion priority is designated to the manufacturing robot.

\appendix

\subsection{Continuous vs Non-continuous Recovery Process}\label{appendix:non-continuous}

\begin{figure}[!t]
    \centering
    \includegraphics[width=0.8\linewidth]{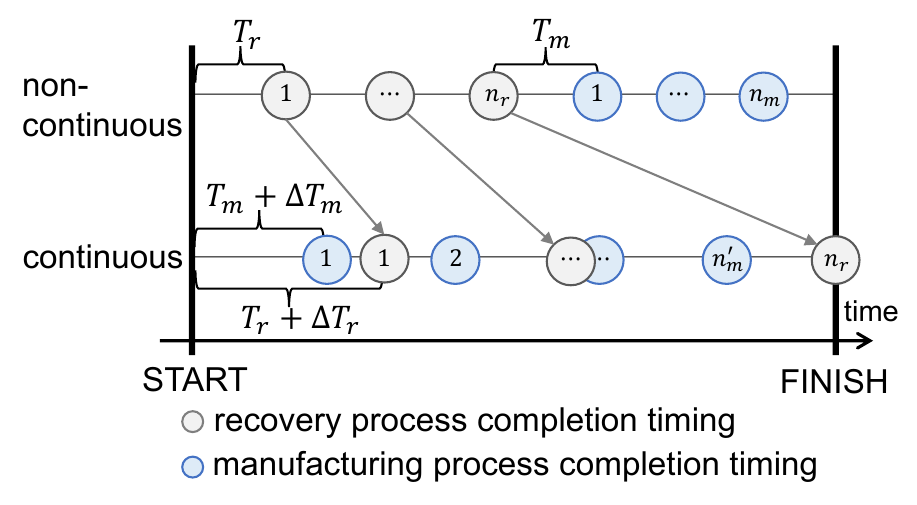}
    \caption{Productivity comparison between the conventional non-continuous process (upper row) and our proposed continuous process (lower row).}
    \label{fig:non-continuous}
\end{figure}

Although the proposed methodology offers advantages, its application may not consistently enhance productivity in the manufacturing recovery process. Certain scenarios reveal that the proposed continuous process (depicted in the right column of Fig.~\ref{fig:background}) may result in lower productivity compared to the non-continuous process (depicted in the middle column of Fig.\ref{fig:background}), as introduced in Section \ref{sec:intro}. To elucidate these circumstances, we conduct a comparative analysis of completion times in the non-continuous process and our proposed continuous process, as illustrated in Fig.~\ref{fig:non-continuous}.
In our analysis, we designate $n_r = 3$ as the constant number of completed recovery processes. Additionally, we employ $n_m$ and $n'_m$ to denote the number of completed manufacturing processes, signifying productivity, before executing $n_r$ recovery processes in non-continuous and continuous recovery processes, respectively. The times required for recovery and manufacturing in the non-continuous process are denoted as $T_r$ and $T_m$, respectively. Furthermore, $\Delta T_r$ and $\Delta T_m$ represent the increments in time required for each recovery and manufacturing process in the continuous process relative to the non-continuous process.
With reference to Fig.~\ref{fig:non-continuous}, we can establish conditions under which the continuous recovery performance surpasses that of the non-continuous approach:
\begin{align}
    \begin{aligned}
        &\ n_m < n'_m \\
        \Leftrightarrow & \ \frac{n_r(T_r+\Delta T_r)-n_r T_r}{T_m} < \frac{n_r(T_r+\Delta T_r)}{T_m+\Delta T_m} \\
        \Leftrightarrow & \ T_r T_m - \Delta T_r \Delta T_m > 0
    \end{aligned}
\end{align}
This condition implies that if the durations for recovery and manufacturing ($T_r$ and $T_m$) in a non-continuous process surpass the augmented durations ($\Delta T_r$ and $\Delta T_m$) in a continuous process, the continuous recovery approach might yield more favorable outcomes. Stated differently, when the execution of recovery and manufacturing tasks is challenging and the incremental task complexity within our continuous process remains modest, the continuous recovery method would outperform its non-continuous counterpart. Consequently, the implementation of our proposed methodology in practical scenarios necessitates an evaluation to ascertain whether this condition is satisfied after determining the optimal priority function design.

\subsection{Data Collection for Imitation Learning}\label{appendix:data-collection}

\begin{figure}[t]
    \centering
    \begin{minipage}[b]{0.47\hsize}
        \centering
        \includegraphics[clip,width=0.99\linewidth]{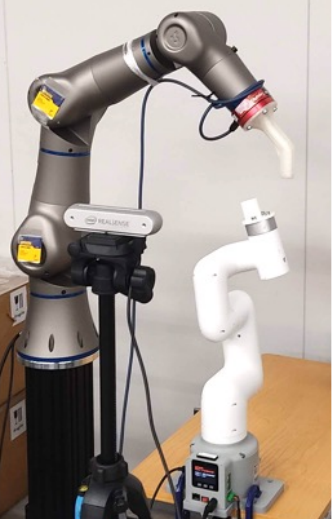}
        \subcaption{Demonstration environment.}
        \label{fig:il_env}
    \end{minipage}
    \begin{minipage}[b]{0.42\hsize}
        \centering
        \includegraphics[clip,width=0.99\linewidth]{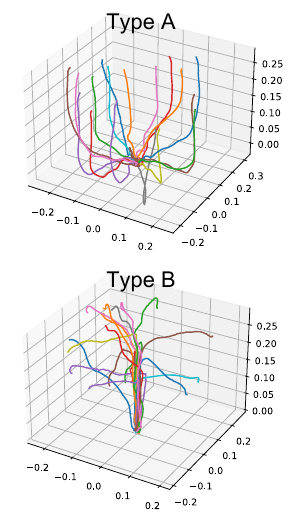}
        \subcaption{Collected trajectories.}
        \label{fig:il_trj}
    \end{minipage}
    \caption{Dataset collection methodology for imitation learning, aimed at simulating manufacturing recovery motions executed by a human operator. (a) The demonstration setup features a compact robot arm that indicates target positions, in conjunction with the teleoperation system detailed in Section \ref{sec:teleoperation_system}. (b) The dataset encompasses two distinct demonstration trajectories: Type A entails an initial vertical adjustment followed by horizontal movement towards the target; Type B commences with positioning directly above the target, succeeded by a descent towards the designated position.}
\end{figure}

For the human recovery demonstration data utilized in ProMPs training, as detailed in Section \ref{sec:promp}, we amassed two distinct categories of human reaching data, denoted as types A and B. This data collection was facilitated by the teleoperation system delineated in Section \ref{sec:teleoperation_system}. Within this setup, an operator manipulates the robot's end-effector to attain a target position. This target, randomly designated, is indicated by a miniature robot arm (myCobot, Elephant Robotics), as depicted in Fig.\,\ref{fig:il_env}.
In the type A approach, the operator first adjusts the vertical position of the controller and subsequently transitions it horizontally to reach the target position. Conversely, in the type B method, the operator first modifies the horizontal orientation of the controller and then proceeds vertically to achieve the target position. Due to constraints on the target robot's reachability, we calibrated both spatial and temporal dimensions of the dataset to align with the requirements of the manufacturing recovery context. Figure \ref{fig:il_trj} visually presents the compiled human reaching data, encompassing 17 trajectories for each demonstration type.

\section*{Acknowledgments}
We thank Dr. Natsuki Yamanobe, Dr. Ixchel G. Ramirez-Alpizar, Dr. Toshio Ueshiba, Dr. Noriaki Ando and Dr. Tamio Tanikawa at AIST for their valuable advice.



\bibliographystyle{IEEEtran}
\bibliography{library}%

\end{document}